\documentclass[11pt]{article}

\usepackage[preprint]{acl}

\usepackage{times}
\usepackage{latexsym}

\usepackage[T1]{fontenc}

\usepackage[utf8]{inputenc}

\usepackage{microtype}

\usepackage{inconsolata}

\usepackage{graphicx}

\usepackage{hyperref}
\usepackage{url}
\usepackage{subfigure}
\usepackage{subcaption}

\usepackage[utf8]{inputenc} 
\usepackage[T1]{fontenc}   
\usepackage{hyperref}       
\usepackage{url}   
\usepackage{booktabs}       
\usepackage{amsfonts}       
\usepackage{nicefrac}       
\usepackage{microtype}      
\usepackage{xcolor}         

\usepackage{lipsum}
\usepackage{wrapfig}
\usepackage{amsmath}
\usepackage{amssymb}
\usepackage{multirow,makecell}
\usepackage{mathtools}
\usepackage{amsthm}
\usepackage[textsize=tiny]{todonotes}

\usepackage{enumitem}
\usepackage{float}
\usepackage[ruled,vlined]{algorithm2e}
\usepackage{xspace}

\newcommand{\method}{LoopMol\xspace}
\newcommand{\methodgp}{LoopMol-GP\xspace}
\newcommand{\methodllm}{LoopMol-LLM\xspace}

\usepackage{multirow}
\usepackage{array}
\usepackage{graphicx}
\usepackage{amsmath,amssymb}
\usepackage{hyperref}
\usepackage{enumitem}
\usepackage{amsfonts}
\usepackage{listings}
\usepackage[most]{tcolorbox}
\usepackage{colortbl}
\definecolor{promptbg}{HTML}{F7FAFC}
\definecolor{promptframe}{HTML}{2F5D7E}
\definecolor{prompttitlebg}{HTML}{2F5D7E}
\definecolor{prompttitlefg}{HTML}{FFFFFF}

\lstdefinestyle{promptstyle}{
    basicstyle=\ttfamily\footnotesize,
    breaklines=true,
breakatwhitespace=false,
columns=fullflexible,
    keepspaces=true,
    showstringspaces=false
}

  \tcbset{
    stageiibox/.style={
      width=\columnwidth,
      top=6pt,
      bottom=6pt,
      left=8pt,
      right=8pt,
      boxsep=2pt,
      arc=2pt,
      colback=black!2!white,
      colframe=black!55,
      coltitle=white,
      colbacktitle=black!80,
      fonttitle=\bfseries,
      enhanced,
      breakable,
      attach boxed title to top left={yshift=-0.08in,xshift=0.12in},
      boxed title style={
        boxrule=0pt,
        arc=1pt,
        colframe=white,
        colback=black!80,
      },
    }
  }
  \newtcolorbox{AIbox}[2][]{stageiibox,title=#2,#1}

\title{
Closed-Loop Molecular Inverse Design with Semantic LLM Surrogates
}

\author{Yaoyao Xu\textsuperscript{1,2}, Xinjian Zhao\textsuperscript{1,2}, Xiaozhuang Song\textsuperscript{1,2}, Lei Bai\textsuperscript{2}, Tianshu Yu\textsuperscript{1,2} \\
  \textsuperscript{1} School of Data Science, The Chinese University of Hong Kong, Shenzhen \\
  \textsuperscript{2} Shanghai Artificial Intelligence Laboratory \\
   \\}

\begin{document}
\maketitle

\begin{abstract}
Practical molecular inverse design is rarely a one-shot generation problem; it often involves enriching a pool of molecules that match a desired property profile under a limited oracle budget. Bayesian optimization (BO) offers a natural framework for this setting, with standard Gaussian-process (GP) surrogates operating in continuous embedding spaces.
We propose \textbf{\method}, a closed-loop framework whose large language model (LLM) surrogate reasons directly over task instructions, SMILES-level optimization histories, and oracle feedback. At each iteration, the surrogate selects informative reference molecules under an exploration--exploitation principle, optionally with a concise guidance sentence. These outputs condition the next generation round and yield an inspectable optimization trace. Experiments on six drug and material tasks across three Llamole backbones show that the LLM-based variant improves over one-shot generation and is competitive with or stronger than GP-based BO. They also reveal a domain-dependent interface: reference-only conditioning performs better on binary drug tasks, while adding a concise guidance sentence generally benefits continuous material tasks.
\end{abstract}

\section{Introduction}
\label{sec:intro}

Molecular inverse design aims to construct molecules that satisfy desired
biological or physicochemical property profiles, and is a core problem in drug
discovery and materials science~\citep{sanchez2018inverse,lu2022inverse,butler2018machine,zunger2018inverse}.
Beyond the enormous size of chemical space, often estimated to contain around
$10^{60}$ drug-like molecules \cite{reymond2015chemical}, a central challenge is that the useful
region for a given target profile can be small, irregular, and difficult to
reach by unconstrained generation.
For screening applications, an important objective is
\emph{candidate-pool enrichment}: increasing the fraction of generated
molecules that match the desired target under a limited oracle
budget~\cite{brown2019guacamol,polykovskiy2020molecular,huang2021therapeutics}.

Large language models (LLMs) have shown promise in molecular representation, property
reasoning, and conditional generation~\cite{chithrananda2020chemberta,edwards2022translation,edwards2021text2mol,taylor2022galactica,liu2023multi}.
For example, Llamole~\cite{liu2025multimodal} combines an instruction-following
language model with a graph diffusion transformer for molecule generation.
When used in a one-shot setting, such models generate candidates from a target
instruction without incorporating feedback on their outputs.
Iterative generation offers an opportunity to use this feedback: molecules
generated in earlier rounds reveal which structures align with the target and
provide references for further search.

Bayesian optimization (BO) provides a natural approach to sequential search
with black-box feedback~\cite{shahriari2015taking,snoek2012practical,jones1998efficient}.
In molecular design, BO commonly uses a Gaussian-process (GP) surrogate over
latent codes or learned embeddings~\cite{gomez2018automatic,kusner2017grammar,jin2018junction,seeger2004gaussian}.
The surrogate models observed scores, and an acquisition function guides the
next evaluation.
This approach can also guide an LLM through reference molecules: BOPRO
retrieves historical candidates near a BO proposal and uses them to construct
the next generation prompt~\cite{agarwal2025searching}.
These methods use numerical representations of the search history, whereas
molecular histories also contain structures, target descriptions, and textual
context that an LLM can process directly.

LLMs provide a flexible input interface for jointly processing molecular
structures, property scores, and natural-language design requirements.
SMILES strings~\cite{weininger1988smiles} describe the candidate structures,
while oracle scores record their agreement with the target.
This interface allows reference selection to account for structural patterns
in the optimization history together with the target context.
We use an LLM as a \emph{semantic surrogate} to relate these observations to
the design requirements and guide subsequent generation.
The surrogate selects references by considering both their scores and their
structural diversity, and can summarize its analysis in a short guidance
sentence.
It also produces natural-language rationales that can be inspected alongside
the selected molecules and their scores.

We introduce \method, a closed-loop framework for molecular inverse design.
Starting from an initial set of candidate molecules, \method evaluates their
properties and uses the accumulated results to select references for
subsequent generation rounds.
\methodllm uses the semantic LLM surrogate, while \methodgp provides a
GP-based baseline with acquisition-guided reference retrieval within the
same framework. We evaluate both on drug and material design tasks and
examine how reference selection and textual guidance affect performance.
Our contributions are as follows:

\begin{itemize}[leftmargin=*]
  \item \textbf{Closed-loop molecular inverse design.}
  We formulate molecular design as iterative candidate-pool enrichment and
  develop \method to connect generation, oracle evaluation, and feedback
  across optimization rounds. We study GP-based and LLM-based variants
  within this common loop.
  \item \textbf{Semantic LLM surrogate.}
  We propose a surrogate that jointly processes molecular structures, oracle
  scores, and target instructions to select reference molecules and provide
  optional textual guidance. The surrogate also produces natural-language
  rationales that can be inspected alongside the selected references.
  \item \textbf{Empirical evaluation and analysis.}
  We evaluate \method across three Llamole backbones and six tasks.
  The LLM variant outperforms the GP variant in most settings, with the
  clearest gains on drug tasks. We find that reference-only conditioning
  performs better on drug tasks, while textual guidance generally improves
  material design, and analyze how reference selection evolves over rounds.
\end{itemize}

\section{Related Work}
\label{sec:related}
\textbf{Goal-directed molecular design.} Goal-directed molecular design has progressed from rule-based search and
reinforcement learning to deep generative models that optimize molecules in
learned latent spaces or directly in graph space ~\cite{olivecrona2017molecular, gomez2018automatic,jin2018junction,popova2018deep,you2018graph}.
Representative systems include policy-based methods, latent-variable models, and
benchmark-driven optimization frameworks such as GuacaMol and PMO~\cite{brown2019guacamol, gao2022sample}.
Recent work has begun to explore LLMs for instruction-conditioned molecular
generation~\cite{bhattacharya2024large,wang2025survey,liu2025multimodal}. Llamole~\cite{liu2025multimodal}, for example, couples an
instruction-following LLM with a graph diffusion decoder.
Most of this literature either retrains the generator for a target objective or uses the generator in a one-shot manner, leaving iterative oracle-guided conditioning less explored.

\noindent \textbf{LLMs as surrogate models in optimization.}
Recent work has explored LLMs as components within sequential optimization and
BO loops. LLAMBO \cite{liu2024large} integrates LLMs into BO for warm-starting,
surrogate modeling, and candidate sampling from natural-language descriptions of
optimization problems. OPRO \cite{yang2023large} uses LLMs as optimizers that
iteratively propose new solutions conditioned on previously evaluated solutions
and their scores. These works show that language models can serve as flexible
optimization modules, but they focus mainly on hyperparameter, prompt, or
configuration spaces. Molecular inverse design raises an additional challenge: the search objects are discrete chemical structures, and the optimization
signal must be translated back into conditional generation.


\section{Preliminaries}
\label{sec:background}

\textbf{Bayesian Optimization.}
\label{sec:bg_bo}
Bayesian optimization (BO)~\cite{frazier2018tutorial} maximizes a black-box function $f: \mathcal{X} \to \mathbb{R}$ by maintaining a probabilistic surrogate over $f$ and iteratively selecting candidates that maximize an acquisition function. Starting from a prior $p(f)$ that encodes initial beliefs about $f$ (e.g., smoothness through a kernel), the surrogate is updated to a posterior $p(f \mid \mathcal{D}_n)$ after observing data $\mathcal{D}_n = \{(\mathbf{x}_i, y_i)\}_{i=1}^n$, which yields a predictive mean $\mu(\mathbf{x})$ and uncertainty $\sigma(\mathbf{x})$ at any query point. The acquisition function $\alpha(\mathbf{x})$ then trades off exploitation (high $\mu$) against exploration (high $\sigma$) to choose the next evaluation point. With a Gaussian-process surrogate and Expected Improvement (EI) acquisition, this takes the closed form
\begin{equation}
  \begin{aligned}
    \mathrm{EI}(\mathbf{x})
      &= (\mu(\mathbf{x})-f^*)\,\Phi(\gamma) + \sigma(\mathbf{x})\,\phi(\gamma), \\
    \gamma &= \frac{\mu(\mathbf{x}) - f^*}{\sigma(\mathbf{x})}.
  \end{aligned}
  \label{eq:ei}
\end{equation}
where $f^*$ is the best observed value and $\Phi, \phi$ are the standard normal CDF and PDF: the first term rewards high predicted mean (\emph{exploitation}), while the second rewards high uncertainty (\emph{exploration}).

\noindent \textbf{Molecular generator.}
Our framework uses an instruction-conditioned molecular generator
\(\mathcal{M}\) that maps text prompts to candidate molecules.
In this paper, \(\mathcal{M}\) is instantiated by
Llamole~\cite{liu2025multimodal}, which couples an instruction-following
language model with a graph-based molecular decoder.
Given a design prompt, the language model produces a conditioning
representation that guides graph generation, and the sampled graph is decoded
into a SMILES string. Its parameters are held fixed during optimization.

\section{Method}
\label{sec:method}

\begin{figure*}[t]
  \centering
  \includegraphics[width=\linewidth]{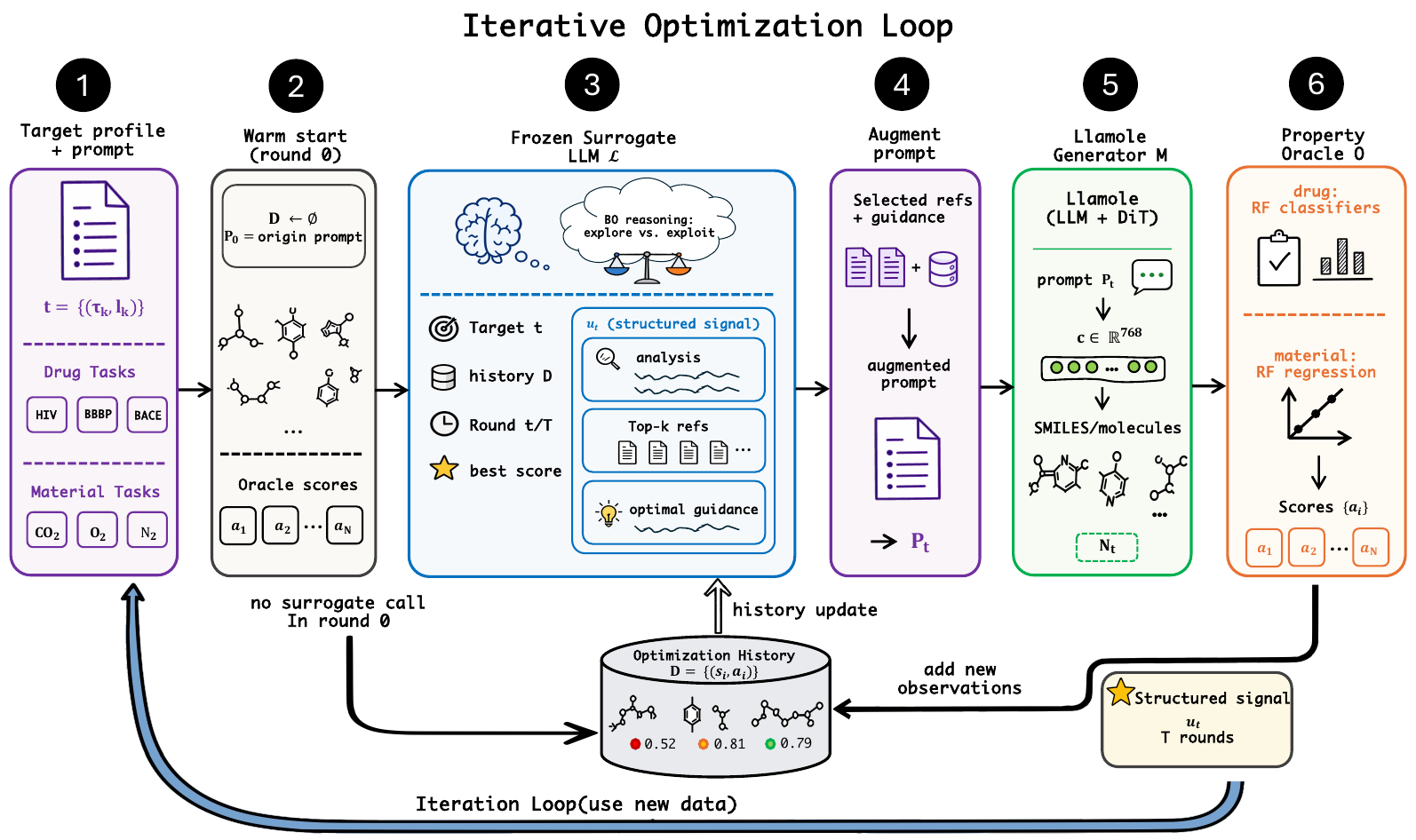}
  \caption{\textbf{Overview of \methodllm.}
  A frozen molecular generator $\mathcal{M}$ and a property oracle
  $\mathcal{O}_\tau$ drive a closed loop of $T$ rounds, with each round's
  SMILES scored and appended to the history $\mathcal{D}$. From round~1 onward,
  a frozen surrogate LLM $\mathcal{L}$ reads the task, history, round index,
  and best score, and emits a structured signal $u_t$ (trajectory analysis,
  top-$k$ references, optional one-sentence guidance) under an exploration
  and exploitation principle, which is parsed into the next-round prompt
  $p_t$. The generator and oracle are held fixed throughout.} 
  \label{fig:overview}
\end{figure*}

\method consists of a molecular generator \(\mathcal{M}\), a task
oracle \(\mathcal{O}_\tau\), and a surrogate decision module \(\mathcal{L}\).
At each round, the generator produces candidate molecules, the oracle scores
them, and the surrogate uses the accumulated history to construct the next
prompt.
We formalize this loop below through the optimization objective, the
task-specific alignment score, and the surrogate modules.

\subsection{Problem Formulation}
\label{sec:method_problem}

With this closed-loop structure in place, each inverse-design instance begins with a natural-language instruction $x$. This instruction specifies the desired molecular design problem, including property requirements, structural preferences, and synthesis-related context. We denote the generator prompt at round $t$ by $p_t$, with $p_0=x$ at warm start.
The generated batch $\mathcal{S}_t$ is:
\begin{equation}
\begin{gathered}
\mathcal{S}_t=\{s_i^{(t)}\}_{i=1}^{N_t}, \\
s_i^{(t)} \sim \mathcal{M}(p_t).
\end{gathered}
\end{equation}

Each run optimizes a \emph{single} oracle task $\tau$.
For drug design, $\tau \in \{\textsc{HIV}, \textsc{BBBP}, \textsc{BACE}\}$ and
the target is a desired binary label $y_\tau^\star \in \{0,1\}$.
For material design, $\tau \in \{\mathrm{CO}_2,\mathrm{O}_2,\mathrm{N}_2\}$ and
the target is a desired continuous permeability value
$y_\tau^\star \in \mathbb{R}_{\ge 0}$.
The design instruction may contain additional conditions beyond $\tau$,
but the optimization objective for a run is always defined with respect to this
single chosen task.

We use a task-specific oracle $\mathcal{O}_\tau$ to evaluate each generated
molecule.
For a molecule $s$, the oracle returns a raw task prediction
$\hat{y}_\tau(s)$:
a class probability for drug tasks and a regression value for material tasks.
Because these raw predictions live on different scales and have different
desiderata, we convert them into a unified alignment score
\(
a_\tau(s; y_\tau^\star)\in[0,1]
\),
where larger values always mean better agreement with the target.
The optimization history after round $t$ is therefore:
\begin{equation}
\begin{aligned}
\mathcal{D}_t
  &= \bigcup_{\nu=0}^{t}\{(s_i^{(\nu)}, a_i^{(\nu)})\}_{i=1}^{N_\nu}, \\
a_i^{(\nu)}
  &= a_\tau(s_i^{(\nu)};y_\tau^\star).
\end{aligned}
\end{equation}
Let $\mathcal{P}=\bigcup_{t=0}^{T-1}\mathcal{S}_t$
denote the full population of molecules generated over the multi-round
trajectory.
To obtain a pool of promising candidates for downstream screening,
we aim to increase the mean alignment score over all generated molecules:
\begin{equation}
\bar{a}_\tau(\mathcal{P}; y_\tau^\star)
= \frac{1}{|\mathcal{P}|}\sum_{s \in \mathcal{P}} a_\tau(s; y_\tau^\star).
\end{equation}
Our evaluation likewise considers candidates from all rounds, using the
task-specific metrics described in Section~\ref{sec:exp_setup}.

\subsection{Task-Specific Alignment Score}
\label{sec:method_oracle}

We use a task-specific alignment score \(a_\tau(s; y_\tau^\star)\in[0,1]\) so
that larger values always mean better agreement with the target, regardless of
whether the oracle is a classifier or a regressor.
For drug tasks, the oracle outputs the probability
\(
\hat{y}_\tau(s)=P_\tau(y=1\mid s)
\).
The target label is $y_\tau^\star \in \{0,1\}$, so we define
\begin{equation}
  a_\tau(s; y_\tau^\star)=
  \begin{cases}
    \hat{y}_\tau(s)       & \text{if } y_\tau^\star = 1, \\
    1-\hat{y}_\tau(s)     & \text{if } y_\tau^\star = 0.
  \end{cases}
  \label{eq:alignment_k}
\end{equation}
This converts the raw oracle output into a score that is always maximized by
better target alignment.

For material tasks, the oracle outputs a continuous permeability
prediction $\hat{y}_\tau(s)$.
We optimize one gas permeability at a time and compare the prediction and the
target in log space. Writing $\ell(y)=\log_{10}(1+y)$, we define
\begin{equation}
  a_\tau(s; y_\tau^\star)=
  \exp\!\left(
  -\frac{\left|\ell(\hat{y}_\tau(s))-\ell(y_\tau^\star)\right|}
        {\sigma_\tau}
  \right),
\end{equation}
where $\sigma_\tau>0$ is a task-specific scale parameter.
This score equals 1 when the prediction matches the target exactly and decays
smoothly as the prediction moves away from the target.

\noindent \textbf{Closed-loop optimization.}
\method runs an independent optimization loop for each test instance over
$T$ rounds.
Round 0 is a warm-start step that uses only the original instruction $x$.
For rounds $t \ge 1$, a surrogate $\mathcal{L}$ reads the current
history $\mathcal{D}_{t-1}$ and produces a decision that conditions the next
generator call.
Concretely, the surrogate selects a top-$k$ reference set
$R_t \subseteq \mathcal{D}_{t-1}$,
and may additionally produce a short conditioning text $g_t$.
These outputs are converted into an augmented generator prompt
\begin{equation}
p_t = \textsc{AugmentPrompt}(x, R_t, g_t),
\end{equation}
which is passed to $\mathcal{M}$ for the next batch of molecule generation.

Reference selection connects accumulated observations to the next search
decision, following the exploration--exploitation principle of BO.
The selected molecules provide examples for refining promising structures
or exploring alternatives, while the generator proposes new candidates
conditioned on these references. Oracle evaluations of the resulting batch
are then added to the history, allowing subsequent selections to account
for the outcomes of earlier decisions.
Section~\ref{sec:method_surrogate} describes the two variants,
\methodllm and \methodgp, which differ in how they use the history to select
references and guide generation.

\subsection{Surrogate Variants}
\label{sec:method_surrogate}

We first describe the semantic LLM surrogate in \methodllm, then the
GP-based reference selection used by \methodgp.

\subsubsection{Semantic LLM Surrogate}
\label{sec:method_llm_surrogate}

\methodllm uses an instruction-following LLM as a semantic surrogate over the
optimization history. The LLM is called once per round and its parameters
remain fixed.
The surrogate input contains the original instruction $x$, the chosen oracle
task $\tau$, the target value $y_\tau^\star$, the available history
$\mathcal{D}_{t-1}$, and the current round.
The prompt asks the LLM to compare structural patterns among candidates with
different scores and select references that balance target alignment and
structural diversity. It emphasizes exploration in early rounds and
exploitation in later rounds.
The LLM makes this trade-off through in-context reasoning over the history,
without explicitly computing an acquisition function.
The surrogate output is a structured signal
\begin{equation}
u_t = \bigl(r_t,\; R_t,\; g_t\bigr),
\end{equation}
where $r_t$ is a natural-language analysis of the current trajectory,
$R_t$ is the selected top-$k$ reference set drawn from the history, and
$g_t$ is an optional one-sentence guidance field.
The output follows a structured format containing the analysis and reference
selection, together with guidance when enabled. The selected references and
optional guidance form the next generator prompt, while the analysis provides
a readable account of the selection.
We consider two prompt interfaces for the augmentation step:
\emph{Top-$k$ only}, which appends $R_t$ and their alignment scores after the
design instruction, and \emph{Top-$k$ + guidance}, which additionally includes
$g_t$ as one concise summary sentence.
The full prompt template is in Appendix~\ref{app:prompts}.

The LLM variant also admits a conceptual analogy to BO: pretrained knowledge
provides a prior, conditioning on the observed history resembles a posterior
update, and reference selection serves an acquisition-like role.
This analogy motivates the design; the LLM does not explicitly represent a
posterior distribution.
Appendix~\ref{app:bo_mapping} discusses this connection.

\subsubsection{GP-Based Reference Selection}
\label{sec:method_gp_surrogate}

\methodgp uses a Gaussian process to model the observed alignment scores
and guide reference selection.
Following BOPRO~\cite{agarwal2025searching}, it retrieves historical molecules
near an acquisition proposal and uses them as references for subsequent
generation.
The GP is fitted on the same alignment scores used by the LLM variant, and
operates in a PCA-compressed space derived from Llamole's conditioning
embeddings.
For each generator input, Llamole produces a 768-dimensional DiT conditioning
embedding $\mathbf{c} \in \mathbb{R}^{768}$.
These embeddings are reduced to $r$ dimensions via truncated SVD, fitted once
on the first batch and then frozen:
\[
\mathbf{z} = \mathbf{V}^\top(\mathbf{c} - \bar{\mathbf{c}})
\in \mathbb{R}^r.
\]
A Mat\'ern-$\nicefrac{5}{2}$ kernel with ARD and $\mathrm{Gamma}(4,2)$ priors
on length scales is fitted via marginal log-likelihood maximization~\cite{snoek2012practical}.
LogEI is then optimized in the PCA
space~\cite{ament2023unexpected,balandat2020botorch} to obtain a proposal
embedding, and the top-$k$ historical molecules nearest this proposal form the
reference set $R_t$.
This variant produces no guidance text $g_t$ and serves as the classical BO
reference under the same prompt-update loop.
\section{Experiments}
\label{sec:experiments}

\subsection{Setup}
\label{sec:exp_setup}

\textbf{Datasets and evaluation metrics.}
We evaluate on the MolQA test set~\cite{liu2025multimodal}, covering two task families: drug design with binary targets over HIV, BBBP, and BACE, and material design with continuous targets over CO$_2$, O$_2$, and N$_2$.
In both domains, each run optimizes a single oracle task, while the original instruction may still include additional design constraints.
We use trajectory-level metrics over \emph{all generated SMILES in all optimization rounds}.
For Drug, we report mean AUC computed from oracle
probabilities against ground-truth labels; invalid SMILES are counted and penalized
as wrong predictions in the AUC evaluation pipeline.
For Material, we report MAE in log$_{10}$ space between oracle predictions and ground-truth targets; invalid SMILES are skipped in MAE computation.
Therefore, for each material metric, we additionally report the invalid ratio (inv.\%) to reflect generation validity.

\noindent \textbf{Experimental details.}
We use the pretrained property oracles provided by Llamole. Drug properties are evaluated by random-forest classifiers on ECFP4 fingerprints~\cite{rogers2010extended}, and material properties are evaluated by random-forest regressors~\cite{rogers2010extended, gao2022sample}.
The surrogate-to-generator interface is configured by domain. For drug tasks, the next-round input is the original instruction followed by the surrogate's selected top-$k$ molecules and their scores.
For material tasks, the same top-$k$ references are augmented with one concise surrogate summary sentence.  The Llamole generator and the
surrogate LLM use the same language-model backbone(Llama-3.1-8B~\cite{grattafiori2024llama}, Mistral-7B~\cite{DBLP:journals/corr/abs-2310-06825}, and Qwen2-7B~\cite{DBLP:journals/corr/abs-2407-10671}), while serving different roles: graph-decoder-based molecule generation and history-conditioned surrogate decision making, respectively. 
More details can be found in Appendix~\ref{app:hyperparams}.

\noindent \textbf{Compared methods.} \textbf{Llamole-OneShot} generates candidates
without iterative refinement. We also include two external one-shot LLMs,
InternS1-mini~\cite{bai2025intern} and Qwen3.5-27B~\cite{qwen35blog}, prompted
directly for molecular design without the Llamole graph decoder.
Among iterative strategies, \textbf{Llamole + Random} draws references
uniformly from the history. \textbf{\methodgp} selects references via BO in a
PCA-compressed embedding space. \textbf{\methodllm} uses the semantic LLM
surrogate to select references and provide optional guidance through the
domain-specific prompt interface described above.

\subsection{Main Results}
\label{sec:exp_main}

\newcommand{\second}[1]{\underline{#1}}

\begin{table*}[ht]
  \centering
  \caption{
  Main results across three generator backbones.
  Drug columns report mean AUC ($\uparrow$) over all generated SMILES.
  Material columns report MAE(log$_{10}$) ($\downarrow$) over valid SMILES only,
  together with the invalid-SMILES ratio (inv.\%) for each target gas.
  BACE and N$_2$, the representative tasks analyzed in the main text,
  are shown first within their respective domains.
  \textbf{Bold} and \underline{underline} indicate the best and second-best
  primary task metrics, respectively, within each backbone group.
  Invalid ratios are reported as auxiliary validity statistics and are not highlighted.
  }
  \label{tab:main}

  \resizebox{\linewidth}{!}{%
  \begin{tabular}{l l ccc cccccc}
    \toprule

    \multirow{2}{*}{\textbf{Backbone}}
    & \multirow{2}{*}{\textbf{Strategy}}
    & \multicolumn{3}{c}{\textbf{Drug AUC $\uparrow$}}
    & \multicolumn{6}{c}{
        \textbf{Material MAE(log$_{10}$) $\downarrow$ + invalid ratio}
      } \\

    \cmidrule(lr){3-5}
    \cmidrule(lr){6-11}

    & & BACE & BBBP & HIV
      & N$_2$ & inv.\%
      & O$_2$ & inv.\%
      & CO$_2$ & inv.\% \\

    \midrule

    External LLM
    & InternS1-mini (OneShot)
    & 0.5499 & 0.5382 & 0.5404
    & 0.9165 & 14.62
    & 0.8023 & 14.48
    & 1.0314 & 15.79 \\

    External LLM
    & Qwen3.5-27B (OneShot)
    & 0.4523 & 0.5526 & 0.4991
    & 0.7800 & 38.54
    & 0.7659 & 41.65
    & 1.1259 & 40.89 \\

    \midrule

    Qwen
    & Llamole-OneShot
    & 0.3261
    & 0.5172
    & 0.4335
    & 0.7700 & 8.95
    & 0.7636 & 17.68
    & \textbf{1.0040} & 5.45 \\

    Qwen
    & Llamole + Random
    & 0.4897
    & 0.5609
    & \second{0.4787}
    & \second{0.7640} & 3.70
    & 0.7631 & 3.24
    & 1.0224 & 3.32 \\

    Qwen
    & \methodgp
    & \second{0.4899}
    & \second{0.5643}
    & \textbf{0.4809}
    & 0.7779 & 3.77
    & \textbf{0.7592} & 3.36
    & 1.0163 & 3.44 \\

    Qwen
    & \methodllm
    & \textbf{0.4972}
    & \textbf{0.5668}
    & 0.4760
    & \textbf{0.7631} & 3.47
    & \second{0.7594} & 4.01
    & \second{1.0101} & 4.37 \\

    \midrule

    Mistral
    & Llamole-OneShot
    & \second{0.6209}
    & 0.6252
    & 0.5644
    & 0.7942 & 4.01
    & 0.7621 & 3.68
    & \second{1.0052} & 3.64 \\

    Mistral
    & Llamole + Random
    & \second{0.6209}
    & \second{0.6253}
    & 0.5643
    & 0.7874 & 1.66
    & 0.7585 & 1.76
    & 1.0060 & 1.93 \\

    Mistral
    & \methodgp
    & 0.6166
    & 0.6229
    & \second{0.5655}
    & \second{0.7864} & 1.54
    & \second{0.7578} & 1.85
    & 1.0058 & 1.58 \\

    Mistral
    & \methodllm
    & \textbf{0.6443}
    & \textbf{0.6293}
    & \textbf{0.5687}
    & \textbf{0.7844} & 1.75
    & \textbf{0.7535} & 1.63
    & \textbf{1.0036} & 1.87 \\

    \midrule

    Llama
    & Llamole-OneShot
    & 0.5111
    & 0.6083
    & \second{0.5579}
    & 0.7671 & 5.04
    & 0.7626 & 5.03
    & 1.0070 & 3.96 \\

    Llama
    & Llamole + Random
    & 0.5638
    & 0.6064
    & 0.5332
    & 0.7660 & 2.59
    & 0.7498 & 2.46
    & 1.0009 & 2.63 \\

    Llama
    & \methodgp
    & \second{0.5707}
    & \second{0.6127}
    & 0.5512
    & \second{0.7646} & 2.66
    & \textbf{0.7473} & 2.37
    & \textbf{0.9941} & 2.45 \\

    Llama
    & \methodllm
    & \textbf{0.6040}
    & \textbf{0.6213}
    & \textbf{0.5653}
    & \textbf{0.7603} & 2.22
    & \second{0.7496} & 2.20
    & \second{0.9967} & 2.36 \\

    \bottomrule
  \end{tabular}
  }
\end{table*}

Table~\ref{tab:main} reveals a consistent picture across all three generator backbones. Closed-loop refinement is broadly beneficial: both random reference selection and GP-BO improve over Llamole-OneShot, confirming that iterative oracle feedback adds value regardless of how the surrogate is implemented. Among iterative strategies, \method matches or surpasses GP-BO in nearly every setting; the advantage is most pronounced on drug tasks, where \method achieves the best AUC on all three targets under both Mistral and Llama. Importantly, this improvement is not tied to a specific generator: the same LLM surrogate mechanism transfers without modification across Qwen, Mistral, and
Llama backbones. Frontier models prompted without graph-constrained decoding (InternS1-mini, Qwen3.5-27B) show high invalid-SMILES rates and are generally outperformed by iterative Llamole strategies, underscoring the role of structured generation in producing chemically valid candidates. The choice of domain-specific interface is not cosmetic: drug runs use the
reference-only template while material runs use the reference-plus-guidance template, and their relative merits are examined directly in the ablation below.

\subsection{Ablation Studies}
\label{sec:exp_ablation}

Beyond reference selection, the \method surrogate can emit a one-sentence
guidance field that synthesizes the optimization trajectory into an explicit
design direction, serving as an interpretable intermediate of the search
process.
Table~\ref{tab:ablation_prompt} isolates the contribution of this field by
comparing \textbf{Top-$k$ only} against \textbf{Top-$k$ + guidance} across
all three backbones.
Guidance improves eight of the nine material settings, while reference-only
conditioning performs better in all nine drug settings.
For binary drug targets, high-scoring reference molecules may already supply a
sufficiently crisp structural signal, leaving little room for the guidance
sentence to contribute further.
For continuous material targets, where the oracle signal is softer and more
graded, the guidance sentence may help compress trajectory-level patterns into
an actionable direction that references alone cannot readily encode.
The preferred interface is largely consistent within each domain across
Qwen, Mistral, and Llama.

\begin{table*}[ht]
  \centering
  \caption{Prompt-interface ablation for \method.
  Rows compare the two prompt interfaces under each generator backbone.
  Drug tasks report mean AUC, and material tasks report
  MAE(log$_{10}$).}
  \label{tab:ablation_prompt}
  \resizebox{0.74\linewidth}{!}{%
  \begin{tabular}{llcccccc}
    \toprule
    \textbf{Backbone} & \textbf{Prompt interface}
    & \textbf{HIV}
    & \textbf{BBBP}
    & \textbf{BACE}
    & \textbf{CO$_2$}
    & \textbf{O$_2$}
    & \textbf{N$_2$} \\
    \midrule
    Qwen & Top-$k$ only & \textbf{0.4760} & \textbf{0.5668} & \textbf{0.4972} & 1.0179 & 0.7656  & 0.7722 \\
    Qwen & Top-$k$ + guidance & 0.4670 & 0.5525 & 0.4859  & \textbf{1.0101}  & \textbf{0.7594} & \textbf{0.7631} \\
    \midrule
    Mistral & Top-$k$ only & \textbf{0.5687} & \textbf{0.6293} & \textbf{0.6443} & 1.0084 & 0.7546 & 0.7875 \\
    Mistral & Top-$k$ + guidance & 0.5521 & 0.6108 & 0.6027 & \textbf{1.0036} & \textbf{0.7535} & \textbf{0.7844} \\
    \midrule
    Llama & Top-$k$ only & \textbf{0.5653} & \textbf{0.6213} & \textbf{0.6040} & 1.0062  & 0.7530 & \textbf{0.7306} \\
    Llama & Top-$k$ + guidance & 0.5387 & 0.6190 & 0.5697 & \textbf{0.9967} & \textbf{0.7496} & 0.7603  \\
    \bottomrule
  \end{tabular}
  }
\end{table*}

\subsection{Analysis}
\label{sec:exp_analysis}

Beyond the performance in Table~\ref{tab:main}, we examine three questions: whether \method improves the search trajectory over rounds, whether the surrogate exhibits genuine exploration--exploitation behavior, and where in chemical space the search budget is allocated.

\begin{figure*}[ht]
    \centering
    \subfigure[BACE Convergence]{
        \includegraphics[width=0.23\linewidth]{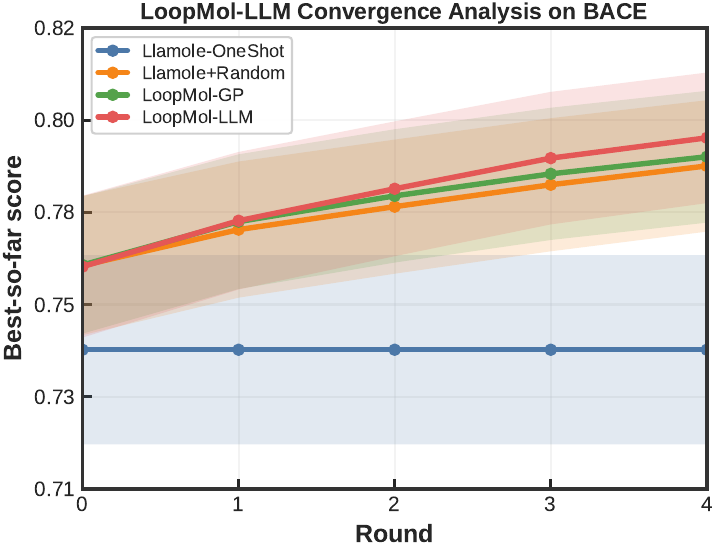}
    }
    \subfigure[BACE Threshold (0.7)]{
        \includegraphics[width=0.23\linewidth]{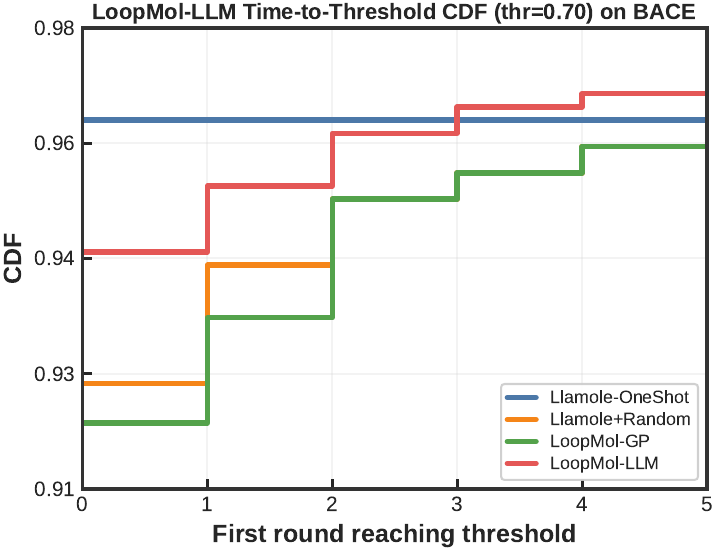}    }
    \subfigure[N$_2$ Convergence]{
        \includegraphics[width=0.23\linewidth]{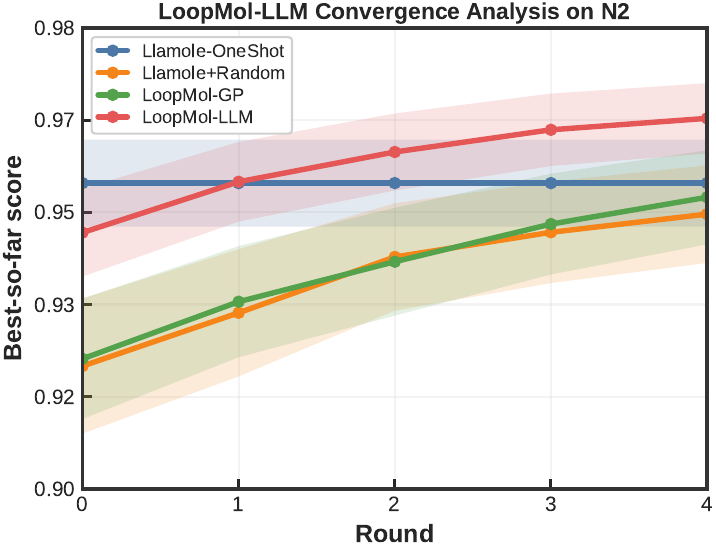}
    }
    \subfigure[N$_2$ Threshold (0.6)]{
        \includegraphics[width=0.23\linewidth]{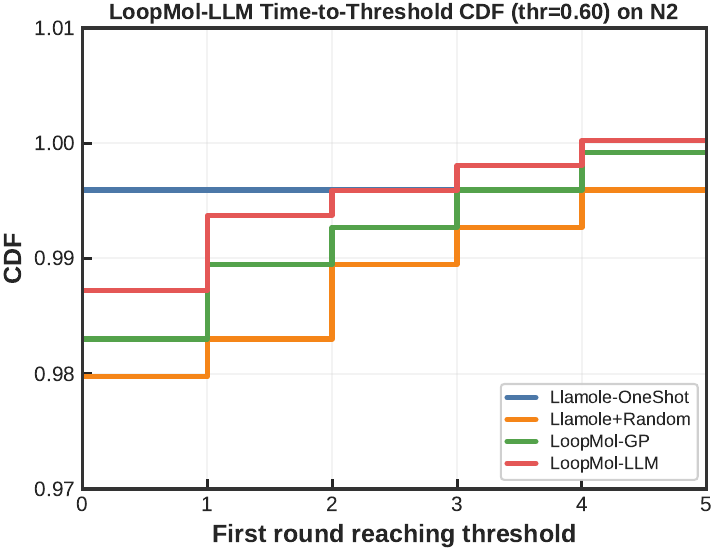}
    }
    \vspace{-3.9mm}\caption{\method discovers high-scoring molecules earlier and more reliably
    than all baselines on representative drug (BACE) and material (N$_2$) tasks.
    Left: mean best-so-far score as a function of round, with 95\% confidence
    intervals across test instances.
    Right: cumulative fraction of samples whose best-so-far score has crossed a
    fixed threshold by each round. Both views show that closed-loop refinement is beneficial, and that
    \method benefits most from iterative feedback.} 
    \label{fig:analysis_convergence}
\end{figure*}

\noindent \textbf{Convergence analysis.}
The closed-loop framing is valuable if oracle feedback materially shapes later rounds.
For each test instance, we track the best-so-far score as a running maximum over rounds, and the cumulative fraction of samples whose best-so-far score has crossed a fixed quality threshold; the first measures search frontier
quality, the second measures how quickly high-scoring molecules are discovered. Figure~\ref{fig:analysis_convergence} shows results on BACE and N$_2$ as representative drug and material tasks. Two patterns emerge consistently: all iterative methods improve as rounds accumulate, confirming that closed-loop refinement is broadly useful; and \method improves more strongly than the baselines, discovering high-scoring molecules earlier and more reliably. Appendix~\ref{app:extra_convergence} provides the visualization for additional tasks.

\noindent \textbf{Effect of iteration rounds.}
We also vary the number of rounds from 3 to 11 on the Llama backbone,
generating 10 molecules per round
(Figure~\ref{fig:round_impact_bace_o2}, Appendix~\ref{app:round_impact}).
BACE AUC increases from 0.5831 to 0.6269, while O$_2$
MAE(log$_{10}$) decreases from 0.7482 to 0.7431.
\method outperforms one-shot generation on both tasks at every tested
round count. These results extend the trajectory analysis: additional
rounds can improve the quality of the full generated population, not
just the best molecule found during search.

\begin{figure*}[h]
    \centering
    \subfigure[BACE]{
        \includegraphics[width=0.23\textwidth]{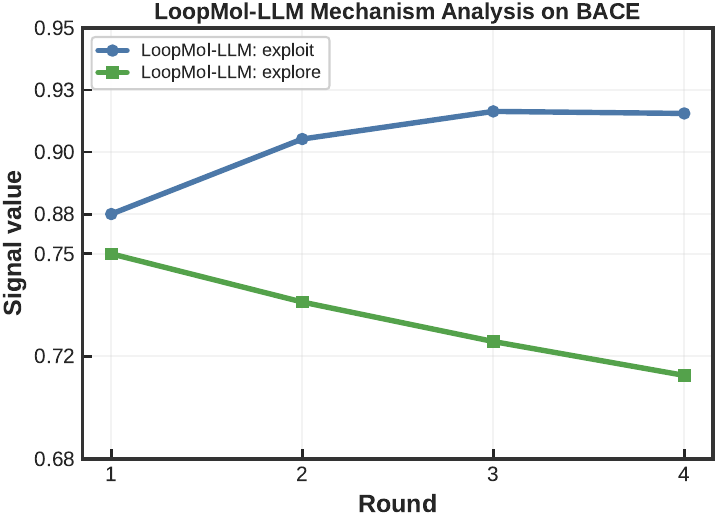}
    }
    \subfigure[BBBP]{
        \includegraphics[width=0.23\textwidth]{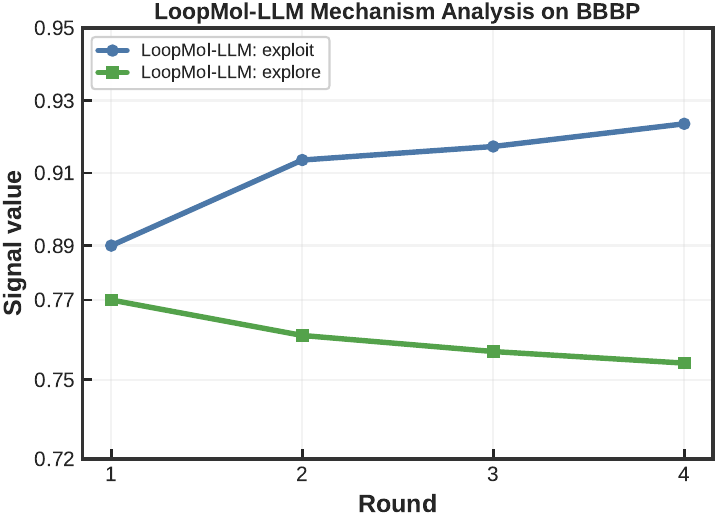}
    }
    \subfigure[N$_2$]{
        \includegraphics[width=0.23\textwidth]{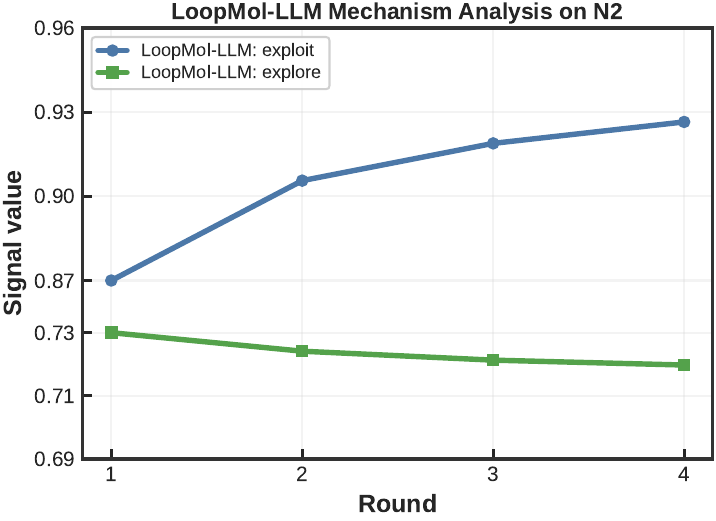}
    }
    \subfigure[O$_2$]{
        \includegraphics[width=0.23\textwidth]{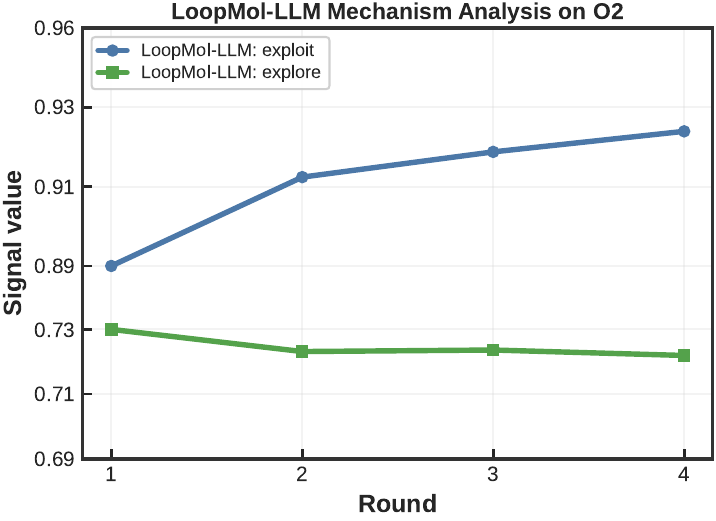}
    } \vspace{-3.9mm}
    \caption{Mechanism analysis of \method across four tasks.
    Each panel summarizes two complementary signals of surrogate behavior:
    the score quantile of selected references within the current history
    (exploitation) and their mean Tanimoto distance to the current top-history
    set (exploration). The shared trend across all four tasks indicates a
    consistent transition from early exploration to later exploitation.} \vspace{-1.9mm}
    \label{fig:analysis_mechanism}
\end{figure*}

\noindent \textbf{BO-style surrogate behavior.}
Better convergence could arise from simple greedy exploitation, reusing only the current top-scoring molecules each round. To examine whether the surrogate instead balances exploration and exploitation, we track two complementary quantities per round: the score quantile of the selected molecules within the available history (higher values indicate stronger exploitation, as the surrogate favors molecules near the top of the observed score distribution) and their mean Tanimoto distance to the current top-history
set (larger values indicate broader exploration).
Figure~\ref{fig:analysis_mechanism} shows a consistent pattern across BACE,
BBBP, N$_2$, and O$_2$: the score quantile increases over rounds while the
distance decreases, a transition from broad early exploration to focused
exploitation as history grows.
This pattern appears across both task families and is consistent
with the intended exploration--exploitation schedule.

\noindent \textbf{Case study.} The analyses above characterize the surrogate's behavior statistically; we now
look inside a concrete trace to show what form its reasoning takes.
We include one compact case study from the N$_2$ task using the Llama backbone.
For this sample, the best score improves from $0.620$ at round 0 to $0.866$ by
round $3$.
The surrogate describes structural patterns in the history and explains its
reference choices, providing a readable account of the search alongside the
observed scores.
The full surrogate prompt template is given in Appendix~\ref{app:prompts};
a condensed version of the actual round-2 prompt and the corresponding
response is provided in Appendix~\ref{app:n2_case}.

At round 2, the surrogate identifies heteroatom-containing structures among
the higher-scoring candidates and selects molecules 10, 36, and 39.
Its guidance combines refinement of high-scoring fluorinated structures
with exploration of alternative heteroatoms. This gives a concrete form to
the exploration--exploitation principle: retain promising structural features
while proposing variations for subsequent evaluation. The selected molecules
supply examples for the next generation round, and the guidance specifies
which features to refine or vary. The oracle then evaluates the resulting
candidates, adding new observations for subsequent reference selection.
As later rounds accumulate stronger candidates, the surrogate shifts its
emphasis toward simpler and more symmetric motifs, indicating that the guidance
is updated by the observed trajectory rather than copied from a fixed heuristic.
This example illustrates how the surrogate translates observations about candidate structures and scores into reference choices and a direction. Appendix~\ref{app:extra_cases} provides a complementary BACE hard case with a full multi-round trace.

\noindent \textbf{Search budget allocation.}
The convergence and mechanism analyses show that \method searches more effectively and follows a BO-like policy; we ask here whether this translates into a measurable difference in \emph{where} molecules are generated in
chemical space.
We represent all generated molecules by ECFP4 fingerprints, project them to two
dimensions, and apply a shared high-score threshold across methods so that the highlighted points are directly comparable.
Figure~\ref{fig:tsne} in Appendix~\ref{app:extra_latent} shows a representative example on N$_2$: compared with one-shot generation, random prompting, and GP-BO, \method places a larger fraction of its molecules in regions that contain these high-scoring points, while still covering multiple neighborhoods rather than collapsing to a single cluster.
This geometric result complements the trajectory and behavioral analyses: the semantic surrogate not only discovers better molecules faster, but steers generation toward the parts of chemical space where good molecules are found. Additional visualizations for other tasks are in Appendix~\ref{app:extra_latent}.

\section{Conclusion}
\label{sec:conclusion}
We introduced \method, a closed-loop framework for target-aligned molecular
inverse design under limited oracle feedback. The framework treats molecular
optimization as candidate-pool enrichment and replaces the usual numerical
surrogate with a semantic decision module that reads molecular histories
directly and returns next-round references, optionally with concise guidance.
Across drug and material tasks, \method improves over one-shot generation and
is competitive with, or stronger than, GP-based BO baselines across multiple
generator backbones. The main analyses support the same picture: the surrogate
improves search trajectories over rounds, exhibits an exploration-to-exploitation
transition, and steers generation toward regions enriched with high-scoring
candidates. Together, these results suggest that LLM surrogates are a useful
closed-loop alternative to classical BO surrogates for discrete molecular
design.

\section*{Limitations}
\label{limitation}
Our evaluation covers six MolQA tasks and three Llamole backbones, using
learned property predictors as oracles. The generated molecules have not
been synthesized or tested in wet-lab experiments, so their predicted
properties remain to be experimentally verified.
\section*{Ethical Considerations}
\method is intended to support computational research in drug and material
discovery. Molecular design methods can also be misused to search for
harmful compounds. Practical use therefore requires appropriate safety
screening and domain expertise before synthesis or experimental testing.

\bibliography{main}

@article{sanchez2018inverse,
  title={Inverse molecular design using machine learning: Generative models for matter engineering},
  author={Sanchez-Lengeling, Benjamin and Aspuru-Guzik, Al{\'a}n},
  journal={Science},
  volume={361},
  number={6400},
  pages={360--365},
  year={2018},
  publisher={American Association for the Advancement of Science}
}

@article{lu2022inverse,
  title={Inverse design with deep generative models: next step in materials discovery},
  author={Lu, Shuaihua and Zhou, Qionghua and Chen, Xinyu and Song, Zhilong and Wang, Jinlan},
  journal={National science review},
  volume={9},
  number={8},
  pages={nwac111},
  year={2022},
  publisher={Oxford University Press}
}

@article{reymond2015chemical,
  title={The chemical space project},
  author={Reymond, Jean-Louis},
  journal={Accounts of chemical research},
  volume={48},
  number={3},
  pages={722--730},
  year={2015},
  publisher={ACS Publications}
}

@article{brown2019guacamol,
  title={GuacaMol: benchmarking models for de novo molecular design},
  author={Brown, Nathan and Fiscato, Marco and Segler, Marwin HS and Vaucher, Alain C},
  journal={Journal of chemical information and modeling},
  volume={59},
  number={3},
  pages={1096--1108},
  year={2019},
  publisher={ACS Publications}
}

@article{chithrananda2020chemberta,
  title={ChemBERTa: large-scale self-supervised pretraining for molecular property prediction},
  author={Chithrananda, Seyone and Grand, Gabriel and Ramsundar, Bharath},
  journal={arXiv preprint arXiv:2010.09885},
  year={2020}
}

@inproceedings{edwards2022translation,
  title={Translation between molecules and natural language},
  author={Edwards, Carl and Lai, Tuan and Ros, Kevin and Honke, Garrett and Cho, Kyunghyun and Ji, Heng},
  booktitle={Proceedings of the 2022 Conference on Empirical Methods in Natural Language Processing},
  pages={375--413},
  year={2022}
}

@article{taylor2022galactica,
  title={Galactica: A large language model for science},
  author={Taylor, Ross and Kardas, Marcin and Cucurull, Guillem and Scialom, Thomas and Hartshorn, Anthony and Saravia, Elvis and Poulton, Andrew and Kerkez, Viktor and Stojnic, Robert},
  journal={arXiv preprint arXiv:2211.09085},
  year={2022}
}

@inproceedings{liu2025multimodal,
  title={MULTIMODAL LARGE LANGUAGE MODELS FOR INVERSE MOLECULAR DESIGN WITH RETROSYNTHETIC PLANNING},
  author={Liu, Gang and Sun, Michael and Matusik, Wojciech and Jiang, Meng and Chen, Jie},
  booktitle={International Conference on Learning Representations},
  year={2025}
}

@article{shahriari2015taking,
  title={Taking the human out of the loop: A review of Bayesian optimization},
  author={Shahriari, Bobak and Swersky, Kevin and Wang, Ziyu and Adams, Ryan P and De Freitas, Nando},
  journal={Proceedings of the IEEE},
  volume={104},
  number={1},
  pages={148--175},
  year={2015},
  publisher={IEEE}
}

@article{gomez2018automatic,
  title={Automatic chemical design using a data-driven continuous representation of molecules},
  author={G{\'o}mez-Bombarelli, Rafael and Wei, Jennifer N and Duvenaud, David and Hern{\'a}ndez-Lobato, Jos{\'e} Miguel and S{\'a}nchez-Lengeling, Benjam{\'\i}n and Sheberla, Dennis and Aguilera-Iparraguirre, Jorge and Hirzel, Timothy D and Adams, Ryan P and Aspuru-Guzik, Al{\'a}n},
  journal={ACS central science},
  volume={4},
  number={2},
  pages={268--276},
  year={2018},
  publisher={ACS Publications}
}

@inproceedings{kusner2017grammar,
  title={Grammar variational autoencoder},
  author={Kusner, Matt J and Paige, Brooks and Hern{\'a}ndez-Lobato, Jos{\'e} Miguel},
  booktitle={International conference on machine learning},
  pages={1945--1954},
  year={2017},
  organization={PMLR}
}

@inproceedings{jin2018junction,
  title={Junction tree variational autoencoder for molecular graph generation},
  author={Jin, Wengong and Barzilay, Regina and Jaakkola, Tommi},
  booktitle={International conference on machine learning},
  pages={2323--2332},
  year={2018},
  organization={PMLR}
}

@article{olivecrona2017molecular,
  title={Molecular de-novo design through deep reinforcement learning},
  author={Olivecrona, Marcus and Blaschke, Thomas and Engkvist, Ola and Chen, Hongming},
  journal={Journal of cheminformatics},
  volume={9},
  number={1},
  pages={48},
  year={2017},
  publisher={Springer}
}

@article{gao2022sample,
  title={Sample efficiency matters: a benchmark for practical molecular optimization},
  author={Gao, Wenhao and Fu, Tianfan and Sun, Jimeng and Coley, Connor},
  journal={Advances in neural information processing systems},
  volume={35},
  pages={21342--21357},
  year={2022}
}

@article{liu2024large,
  title={Large language models to enhance bayesian optimization},
  author={Liu, Tennison and Astorga, Nicol{\'a}s and Seedat, Nabeel and van der Schaar, Mihaela},
  journal={arXiv preprint arXiv:2402.03921},
  year={2024}
}

@inproceedings{yang2023large,
  title={Large language models as optimizers},
  author={Yang, Chengrun and Wang, Xuezhi and Lu, Yifeng and Liu, Hanxiao and Le, Quoc V and Zhou, Denny and Chen, Xinyun},
  booktitle={The Twelfth International Conference on Learning Representations},
  year={2023}
}

@inproceedings{agarwal2025searching,
  title={Searching for optimal solutions with LLMs via bayesian optimization},
  author={Agarwal, Dhruv and Arivazhagan, Manoj Ghuhan and Das, Rajarshi and Swamy, Sandesh and Khosla, Sopan and Gangadharaiah, Rashmi},
  booktitle={The Thirteenth International Conference on Learning Representations},
  year={2025}
}

@article{frazier2018tutorial,
  title={A tutorial on Bayesian optimization},
  author={Frazier, Peter I},
  journal={arXiv preprint arXiv:1807.02811},
  year={2018}
}

@article{weininger1988smiles,
  title={SMILES, a chemical language and information system. 1. Introduction to methodology and encoding rules},
  author={Weininger, David},
  journal={Journal of chemical information and computer sciences},
  volume={28},
  number={1},
  pages={31--36},
  year={1988},
  publisher={ACS Publications}
}

@article{rogers2010extended,
  title={Extended-connectivity fingerprints},
  author={Rogers, David and Hahn, Mathew},
  journal={Journal of chemical information and modeling},
  volume={50},
  number={5},
  pages={742--754},
  year={2010},
  publisher={ACS Publications}
}

@article{ament2023unexpected,
  title={Unexpected improvements to expected improvement for bayesian optimization},
  author={Ament, Sebastian and Daulton, Samuel and Eriksson, David and Balandat, Maximilian and Bakshy, Eytan},
  journal={Advances in neural information processing systems},
  volume={36},
  pages={20577--20612},
  year={2023}
}

@article{balandat2020botorch,
  title={BoTorch: A framework for efficient Monte-Carlo Bayesian optimization},
  author={Balandat, Maximilian and Karrer, Brian and Jiang, Daniel and Daulton, Samuel and Letham, Ben and Wilson, Andrew G and Bakshy, Eytan},
  journal={Advances in neural information processing systems},
  volume={33},
  pages={21524--21538},
  year={2020}
}

@article{butler2018machine,
  title={Machine learning for molecular and materials science},
  author={Butler, Keith T and Davies, Daniel W and Cartwright, Hugh and Isayev, Olexandr and Walsh, Aron},
  journal={Nature},
  volume={559},
  number={7715},
  pages={547--555},
  year={2018},
  publisher={Nature Publishing Group UK London}
}

@article{zunger2018inverse,
  title={Inverse design in search of materials with target functionalities},
  author={Zunger, Alex},
  journal={Nature Reviews Chemistry},
  volume={2},
  number={4},
  pages={0121},
  year={2018},
  publisher={Nature Publishing Group UK London}
}

@article{polykovskiy2020molecular,
  title={Molecular sets (MOSES): a benchmarking platform for molecular generation models},
  author={Polykovskiy, Daniil and Zhebrak, Alexander and Sanchez-Lengeling, Benjamin and Golovanov, Sergey and Tatanov, Oktai and Belyaev, Stanislav and Kurbanov, Rauf and Artamonov, Aleksey and Aladinskiy, Vladimir and Veselov, Mark and others},
  journal={Frontiers in pharmacology},
  volume={11},
  pages={565644},
  year={2020},
  publisher={Frontiers}
}

@article{huang2021therapeutics,
  title={Therapeutics Data Commons: Machine Learning Datasets and Tasks for Drug Discovery and Development},
  author={Huang, Kexin and Fu, Tianfan and Gao, Wenhao and Zhao, Yue and Roohani, Yusuf and Leskovec, Jure and Coley, Connor W and Xiao, Cao and Sun, Jimeng and Zitnik, Marinka},
  journal={Advances in Neural Information Processing Systems},
  year={2021},
  publisher={Neural information processing systems foundation}
}

@inproceedings{edwards2021text2mol,
  title={Text2mol: Cross-modal molecule retrieval with natural language queries},
  author={Edwards, Carl and Zhai, ChengXiang and Ji, Heng},
  booktitle={Proceedings of the 2021 conference on empirical methods in natural language processing},
  pages={595--607},
  year={2021}
}

@article{liu2023multi,
  title={Multi-modal molecule structure--text model for text-based retrieval and editing},
  author={Liu, Shengchao and Nie, Weili and Wang, Chengpeng and Lu, Jiarui and Qiao, Zhuoran and Liu, Ling and Tang, Jian and Xiao, Chaowei and Anandkumar, Animashree},
  journal={Nature Machine Intelligence},
  volume={5},
  number={12},
  pages={1447--1457},
  year={2023},
  publisher={Nature Publishing Group UK London}
}

@article{snoek2012practical,
  title={Practical bayesian optimization of machine learning algorithms},
  author={Snoek, Jasper and Larochelle, Hugo and Adams, Ryan P},
  journal={Advances in neural information processing systems},
  volume={25},
  year={2012}
}

@article{jones1998efficient,
  title={Efficient global optimization of expensive black-box functions},
  author={Jones, Donald R and Schonlau, Matthias and Welch, William J},
  journal={Journal of Global optimization},
  volume={13},
  number={4},
  pages={455--492},
  year={1998},
  publisher={Springer}
}

@article{seeger2004gaussian,
  title={Gaussian processes for machine learning},
  author={Seeger, Matthias},
  journal={International journal of neural systems},
  volume={14},
  number={02},
  pages={69--106},
  year={2004},
  publisher={World Scientific}
}

@article{popova2018deep,
  title={Deep reinforcement learning for de novo drug design},
  author={Popova, Mariya and Isayev, Olexandr and Tropsha, Alexander},
  journal={Science advances},
  volume={4},
  number={7},
  pages={eaap7885},
  year={2018},
  publisher={American Association for the Advancement of Science}
}

@article{you2018graph,
  title={Graph convolutional policy network for goal-directed molecular graph generation},
  author={You, Jiaxuan and Liu, Bowen and Ying, Zhitao and Pande, Vijay and Leskovec, Jure},
  journal={Advances in neural information processing systems},
  volume={31},
  year={2018}
}

@article{bai2025intern,
  title={Intern-s1: A scientific multimodal foundation model},
  author={Bai, Lei and Cai, Zhongrui and Cao, Yuhang and Cao, Maosong and Cao, Weihan and Chen, Chiyu and Chen, Haojiong and Chen, Kai and Chen, Pengcheng and Chen, Ying and others},
  journal={arXiv preprint arXiv:2508.15763},
  year={2025}
}

@misc{qwen35blog,
    title = {Qwen3.5: Accelerating Productivity with Native Multimodal Agents},
    url = {https://qwen.ai/blog?id=qwen3.5},
    author = {Qwen Team},
    month = {February},
    year = {2026}
}

@article{grattafiori2024llama,
  title={The llama 3 herd of models},
  author={Grattafiori, Aaron and Dubey, Abhimanyu and Jauhri, Abhinav and Pandey, Abhinav and Kadian, Abhishek and Al-Dahle, Ahmad and Letman, Aiesha and Mathur, Akhil and Schelten, Alan and Vaughan, Alex and others},
  journal={arXiv preprint arXiv:2407.21783},
  year={2024}
}

@article{wang2025survey,
  title={A survey of large language models for text-guided molecular discovery: from molecule generation to optimization},
  author={Wang, Ziqing and Zhang, Kexin and Zhao, Zihan and Wen, Yibo and Pandey, Abhishek and Liu, Han and Ding, Kaize},
  journal={arXiv preprint arXiv:2505.16094},
  year={2025}
}

@article{bhattacharya2024large,
  title={Large language models as molecular design engines},
  author={Bhattacharya, Debjyoti and Cassady, Harrison J and Hickner, Michael A and Reinhart, Wesley F},
  journal={Journal of Chemical Information and Modeling},
  volume={64},
  number={18},
  pages={7086--7096},
  year={2024},
  publisher={ACS Publications}
}

@article{DBLP:journals/corr/abs-2407-10671,
  author       = {An Yang and
                  Baosong Yang and
                  Binyuan Hui and
                  Bo Zheng and
                  Bowen Yu and
                  Chang Zhou and
                  Chengpeng Li and
                  Chengyuan Li and
                  Dayiheng Liu and
                  Fei Huang and
                  Guanting Dong and
                  Haoran Wei and
                  Huan Lin and
                  Jialong Tang and
                  Jialin Wang and
                  Jian Yang and
                  Jianhong Tu and
                  Jianwei Zhang and
                  Jianxin Ma and
                  Jianxin Yang and
                  Jin Xu and
                  Jingren Zhou and
                  Jinze Bai and
                  Jinzheng He and
                  Junyang Lin and
                  Kai Dang and
                  Keming Lu and
                  Keqin Chen and
                  Kexin Yang and
                  Mei Li and
                  Mingfeng Xue and
                  Na Ni and
                  Pei Zhang and
                  Peng Wang and
                  Ru Peng and
                  Rui Men and
                  Ruize Gao and
                  Runji Lin and
                  Shijie Wang and
                  Shuai Bai and
                  Sinan Tan and
                  Tianhang Zhu and
                  Tianhao Li and
                  Tianyu Liu and
                  Wenbin Ge and
                  Xiaodong Deng and
                  Xiaohuan Zhou and
                  Xingzhang Ren and
                  Xinyu Zhang and
                  Xipin Wei and
                  Xuancheng Ren and
                  Xuejing Liu and
                  Yang Fan and
                  Yang Yao and
                  Yichang Zhang and
                  Yu Wan and
                  Yunfei Chu and
                  Yuqiong Liu and
                  Zeyu Cui and
                  Zhenru Zhang and
                  Zhifang Guo and
                  Zhihao Fan},
  title        = {Qwen2 Technical Report},
  journal      = {CoRR},
  volume       = {abs/2407.10671},
  year         = {2024},
  url          = {https://doi.org/10.48550/arXiv.2407.10671},
  doi          = {10.48550/ARXIV.2407.10671},
  eprinttype   = {arXiv},
  eprint       = {2407.10671},
  bibsource    = {dblp computer science bibliography, https://dblp.org}
}

@article{DBLP:journals/corr/abs-2310-06825,
  author       = {Albert Q. Jiang and
                  Alexandre Sablayrolles and
                  Arthur Mensch and
                  Chris Bamford and
                  Devendra Singh Chaplot and
                  Diego de Las Casas and
                  Florian Bressand and
                  Gianna Lengyel and
                  Guillaume Lample and
                  Lucile Saulnier and
                  L{\'{e}}lio Renard Lavaud and
                  Marie{-}Anne Lachaux and
                  Pierre Stock and
                  Teven Le Scao and
                  Thibaut Lavril and
                  Thomas Wang and
                  Timoth{\'{e}}e Lacroix and
                  William El Sayed},
  title        = {Mistral 7B},
  journal      = {CoRR},
  volume       = {abs/2310.06825},
  year         = {2023},
  url          = {https://doi.org/10.48550/arXiv.2310.06825},
  doi          = {10.48550/ARXIV.2310.06825},
  eprinttype   = {arXiv},
  eprint       = {2310.06825},
  bibsource    = {dblp computer science bibliography, https://dblp.org}
}

\appendix

\section{Algorithm}

Algorithm~\ref{alg:llmbo} summarizes the closed-loop procedure of \method,
where the LLM surrogate uses the accumulated oracle history to update the
generation prompt and iteratively enrich the candidate pool.
\begin{algorithm}[ht]
\caption{\method: single-call \method loop for instruction $x$ and oracle task $\tau$}
\label{alg:llmbo}
\SetKwInOut{Input}{Input}
\SetKwInOut{Output}{Output}
\Input{Instruction $x$, target value $y_\tau^\star$, frozen surrogate LLM $\mathcal{L}$,
       frozen generator $\mathcal{M}$, task oracle $\mathcal{O}_\tau$,
       rounds $T$, warm-start size $N_0$, per-round batch size $N$}
\Output{Generated population $\mathcal{P}$ and trajectory history $\mathcal{D}$}
$\mathcal{D} \gets \emptyset$\;
\For{$t = 0, \ldots, T-1$}{
  \eIf{$t = 0$}{
    $p_t \gets x$\;
    $N_t \gets N_0$\;
  }{
    $u_t \gets \mathcal{L}(\textsc{SurrogatePrompt}(x, \tau, y_\tau^\star, \mathcal{D}, t, T))$\;
    $p_t \gets \textsc{AugmentPrompt}(x, u_t)$\;
    $N_t \gets N$\;
  }
  $\mathcal{S}_t = \{s_i^{(t)}\}_{i=1}^{N_t} \gets \mathcal{M}(p_t)$\;
  $\{a_i^{(t)}\}_{i=1}^{N_t} \gets \mathcal{O}_\tau(\mathcal{S}_t, y_\tau^\star)$\;
  $\mathcal{D} \gets \mathcal{D} \cup \{(s_i^{(t)}, a_i^{(t)})\}_{i=1}^{N_t}$\;
}
$\mathcal{P} \gets \bigcup_{t=0}^{T-1}\mathcal{S}_t$\;
\Return $(\mathcal{P}, \mathcal{D})$\;
\end{algorithm}

\section{BO Perspective and LLM Surrogate Mapping}
\label{app:bo_mapping}
The difference between classical BO and \method lies in the surrogate layer.
In the GP baseline, the search state is represented in a compressed continuous
embedding space, the posterior is updated through kernel conditioning, and the
acquisition rule is optimized numerically.
In \method, these operations are replaced by a large language model that reads
the discrete optimization history directly as molecules and scores.

This replacement preserves the decision logic of BO while changing the form of
the surrogate.
The prior is interpreted as the surrogate LLM's pretrained chemical and
optimization knowledge, the posterior-like state update is realized through
in-context conditioning on the full history, and the acquisition-style step
appears as explicit reference selection under an explore--exploit instruction.
Because the search is expressed in terms of molecules and scores rather than
latent coordinates alone, each round becomes readable in natural language.

This design is motivated by the structure of molecular optimization.
First, the search space is discrete and highly compositional, so useful
structure--property patterns may be difficult to capture with a smooth kernel
in a compressed latent space.
Second, the utility of a reference molecule depends on the whole trajectory and
the target context, not only on pairwise geometric proximity.
Third, the surrogate output can be inspected directly through its analysis,
selected references, and guidance, which makes the optimization policy more
transparent than a purely numerical acquisition value.

\section{Complete Prompt Templates}
\label{app:prompts}

\textbf{Baseline generator prompt interface.}
All iterative baselines keep the original MolQA instruction unchanged and only
modify the additional conditioning block appended after the instruction.
The downstream generator-facing template is:
\begin{AIbox}{Shared Generator Prompt}
\small
\begin{lstlisting}[style=promptstyle]
{original MolQA instruction}

Here are some previously designed molecules for reference:
1. SMILES: {smiles_1} (score: {score_1})
2. SMILES: {smiles_2} (score: {score_2})
...
k. SMILES: {smiles_k} (score: {score_k})

Based on these references, design a new and better molecule.
\end{lstlisting}
\end{AIbox}

The difference across baselines is only how the reference set is chosen:
\texttt{Random} samples references uniformly from the previous history;
\texttt{GP-BO} retrieves the top-$k$ molecules nearest to the GP proposal in
embedding space; and \texttt{\method (top-$k$)} uses the surrogate LLM's
selected references.

\noindent \textbf{\method guidance interface.}
For the guidance variant, the same reference information is followed by one
concise surrogate summary sentence:

\begin{AIbox}{\method Guidance Prompt}
\small
\begin{lstlisting}[style=promptstyle]
{original MolQA instruction}

Optimization guidance:
Use [{smiles_1} ({score_1}), ..., {smiles_k} ({score_k})] as references.
{surrogate guidance sentence}.
Design focus on {round-dependent design focus}.

Based on this guidance, design a new and better molecule.
\end{lstlisting}
\end{AIbox}

\noindent \textbf{Surrogate selection prompt.}
The single-call surrogate prompt has the following structure:

\begin{AIbox}{LLM Surrogate Prompt}
\small
\begin{lstlisting}[style=promptstyle]
You are helping optimize molecular design. Your job is to select {k}
reference molecules from the history below. These will guide the next
round of molecule generation.

=== Task ===
{molqa_instruction}
Target properties: {target_properties}
Property ranges:
  {property_name}: target={target}, dataset range=[{min}, {max}]

=== Optimization History ({n} molecules) ===
[1] {SMILES_1} | score={score_1} ({individual oracle outputs})
[2] {SMILES_2} | score={score_2} ({individual oracle outputs})
...

=== Current State ===
Best score: {best_score} (molecule [{best_index}])
Round: {current_round}/{total_rounds}
Stage: {early / mid / late stage instruction}

=== Selection Principle (Bayesian Optimization) ===
Do NOT simply pick the {k} highest-scoring molecules. Instead, balance:
- EXPLOIT: pick molecules with high scores -- these are good structures to refine.
- EXPLORE: pick molecules with diverse or unusual structures -- even if scores are 
moderate, they may represent promising unexplored directions.

=== Output Format ===
ANALYSIS: ...
SELECTED: [{k} molecule numbers, comma-separated]
\end{lstlisting}
\end{AIbox}

When the guidance interface is enabled, the surrogate is additionally asked to
produce a compact guidance field, which is later converted into the
generator-facing summary sentence.

\section{Implementation Details}
\label{app:implementation}

\textbf{Model families.}
Our experiments use two model families with different roles.
The first family is the Llamole generator, which couples an instruction-tuned
language model with graph-generation modules, and is used in all iterative
methods.
The second family is the external one-shot baselines, which are prompted
directly to produce molecular outputs without the Llamole graph decoder or any
iterative optimization loop.
Table~\ref{tab:model_sizes} summarizes the model backbones used in the study. All experiments were conducted on a single compute node with eight NVIDIA A100 GPUs.

\begin{table*}[ht]
  \centering
  \caption{Language-model backbones used in the study.}
  \label{tab:model_sizes}
  \begin{tabular}{lll}
    \toprule
    \textbf{Role} & \textbf{Model} & \textbf{Nominal size} \\
    \midrule
    Llamole backbone & Qwen2-7B-Instruct & 7B \\
    Llamole backbone & Mistral-7B-Instruct-v0.3 & 7B \\
    Llamole backbone & Llama-3.1-8B-Instruct & 8B \\
    External one-shot baseline & Intern-S1-mini & 8B LM + 0.3B vision encoder \\
    External one-shot baseline & Qwen3.5-27B & 27B \\
    \bottomrule
  \end{tabular}
\end{table*}

Intern-S1-mini is used in text-only mode in our one-shot baseline.

\noindent \textbf{Llamole generator.}
Llamole is the molecular generator used throughout the iterative methods.
It combines an instruction-following backbone LLM with a graph diffusion
transformer (DiT), a graph encoder, and a graph predictor.
Given a natural-language design instruction, the backbone LLM produces a
conditioning representation, which is mapped to a 768-dimensional latent vector
used to steer the DiT.
The DiT then samples a molecular graph that is decoded into a SMILES string.
In the original Llamole formulation, the language-model output can also support
textual explanation and retrosynthetic context.
In our work, however, Llamole is used only as the generator: all optimization
logic is handled externally by either a GP surrogate or an LLM surrogate.  The reference codebase is released under the Apache License 2.0.

\noindent \textbf{Classical GP-BO baseline.}
Our classical BO baseline is inspired by latent-space BO methods such as
BOPRO, but it is adapted to the molecular inverse-design setting considered in
this paper.
The baseline is built around a fixed molecular generator and an explicit
property oracle, so the surrogate is optimized against the same
target-aware alignment score used elsewhere in the paper.
This differs from sequence-level verifier objectives and keeps the optimization
target aligned with the final evaluation metric.

The representation used by the GP is the 768-dimensional DiT conditioning
embedding produced by Llamole.
This choice is important because the embedding directly controls the downstream
graph generator, making it a more faithful search space than a generic text
embedding.
To make GP fitting stable in this high-dimensional space, we first run a
warm-start round that generates 30 molecules, fit PCA on the resulting
embeddings, and optimize thereafter in the fixed PCA-projected space.
The GP itself uses a Mat\'ern-$\nicefrac{5}{2}$ kernel with ARD, a standard
choice in practical GP-BO that imposes weaker smoothness assumptions than the
squared-exponential kernel while allowing different PCA dimensions to have
separate lengthscales.
We use LogEI acquisition, and warm-start GP hyperparameters from the previous
round rather than reinitializing them from scratch.

The GP proposal is not decoded directly into a molecule.
Instead, following BOPRO, it retrieves the top-$k$ nearest historical molecules in the embedding
space, and these molecules are appended to the next Llamole prompt as
references.
This keeps the generator interface matched across iterative methods: all of
them steer Llamole through textual conditioning, while differing only in how
the references are chosen.

\textbf{Property oracles.}
Drug tasks use one random-forest classifier per property, and material tasks
use one random-forest regressor per target gas.
For drug tasks, the oracle score is flipped when the desired label is zero so
that higher alignment always means better agreement with the target.
For material tasks, the regression output is converted into an exponential
alignment score in log space so that lower target error corresponds to higher
optimization score.

\section{Data Overview}
\label{app:data}

\textbf{Task format.}
We use the MolQA test set~\cite{liu2025multimodal} for evaluation.
The MolQA test set contains 9,986 QA pairs for small molecules in drug applications and 750 pairs for polymeric materials. Each MolQA instance contains an instruction, an optional input field, a
reference output, and a property dictionary.
The instruction defines the inverse-design problem in natural language, while
the property dictionary specifies the target profile to be matched.
In drug tasks, these targets are binary labels such as HIV activity, BBBP, and
BACE inhibition.
In material tasks, they are continuous values such as CO$_2$, N$_2$, and O$_2$
permeability, along with auxiliary structural or physical attributes.

\noindent \textbf{Drug and material domains.}
The drug tasks evaluate whether the generated molecule matches a desired
multi-property binary profile.
The material tasks evaluate whether the generated molecule is aligned with a
continuous gas-permeability target.
This distinction matters throughout the paper because the oracle signal is
qualitatively different in the two domains: classification-style supervision in
drug design is relatively crisp, whereas regression-style supervision in
material design is softer and more fine-grained.
This is one of the main reasons we study prompt-interface differences across
domains.

\noindent \textbf{Illustrative examples.}
To make the task format concrete, we include one schematic drug-style example
and one material example below.
The purpose of these examples is only to show the interface seen by the
generator and the surrogate; they are not used as evaluation cases.

\begin{AIbox}[colback=green!2!white,colframe=green!40!black,colbacktitle=green!55!black]{Illustrative Drug Example}
\small
\begin{lstlisting}[style=promptstyle]
[Instruction]
Design a molecule that is active against HIV, permeable to the
blood-brain barrier, and inactive against BACE.

[Target properties]
{HIV: 1, BBBP: 1, BACE: 0}
\end{lstlisting}
\end{AIbox}

\begin{AIbox}[colback=teal!2!white,colframe=teal!45!black,colbacktitle=teal!60!black]{Illustrative Material Example}
\small
\begin{lstlisting}[style=promptstyle]
[Instruction]
What is the optimal molecular design and synthesis route for a
polymer with high CO2 gas permeability and low permeability to
N2 and O2, featuring an aromatic ring and specific functional
groups?

[Target properties]
{CO2: 0.94, N2: 0.0, O2: 0.0, FFV: 0.381, SC: 2.28, SA: 4.21}
\end{lstlisting}
\end{AIbox}

\section{Hyper-Parameter Summary}
\label{app:hyperparams}
All iterative methods share the same common optimization budget unless otherwise
specified. We use $T=5$ optimization rounds, with $N_0=30$ warm-start molecules
at round 0 and $N=10$ molecules for each subsequent round. All methods use
Top-$k$ in-context references with $k=3$. Method-specific hyperparameters are
set only where required: the surrogate LLM for \method, and $r=30$ PCA
dimensions, a Mat\'ern-$\nicefrac{5}{2}$ kernel, and LogEI acquisition for the
GP-BO baseline. The full set of shared and method-specific hyperparameters is
summarized in Table~\ref{tab:hyperparameters}.

\begin{table*}[ht]
  \centering
  \caption{Main hyperparameters used in the experiments.}
  \label{tab:hyperparams}
  \begin{tabular}{ll}
    \toprule
    \textbf{Component} & \textbf{Value} \\
    \midrule
    Optimization rounds $T$ & 5 \\
    Warm-start molecules $N_0$ & 30 \\
    Molecules per later round $N$ & 10 \\
    Top-$k$ references & 3 \\
    Surrogate temperature & 0.3 \\
    DiT conditioning dimension & 768 \\
    GP PCA dimension & 30 \\
    GP kernel & Mat\'ern-$\nicefrac{5}{2}$ with ARD \\
    GP acquisition & LogEI \\
    Material guidance mode & Top-$k$ + one-sentence summary \\
    Drug guidance mode & Top-$k$ only \\
    \bottomrule
  \end{tabular}
  \label{tab:hyperparameters}
\end{table*}

\section{Analysis Protocols}
\label{app:analysis_details}

This section summarizes the quantitative protocols used in the analysis
section.

\paragraph{Convergence metrics.}
For sample $i$ at round $t$, let $s_i^{(t)}$ denote the best score obtained
within that round.
We define the best-so-far trajectory as
\[
b_i^{(t)} = \max_{\tau \le t} s_i^{(\tau)}.
\]
The convergence curve reports the sample mean of $b_i^{(t)}$ together with a
95\% confidence interval across test instances.
To measure sample efficiency, we fix a task-specific threshold $\lambda$ and
define the first hitting time
\[
T_i(\lambda) = \min \{ t : b_i^{(t)} \ge \lambda \}.
\]
The corresponding threshold plot reports the cumulative fraction of samples
whose hitting time is at most $t$.
This figure therefore measures how quickly a method reaches the high-score
region.

\paragraph{Mechanism metrics.}
At round $t$, let $\mathcal{H}_t$ be the set of molecules already observed and
let $\mathcal{S}_t$ be the set of references selected by the surrogate.
We measure exploitation through the empirical score quantile of selected
molecules within the available history:
\[
q_t = \frac{1}{|\mathcal{S}_t|}\sum_{x \in \mathcal{S}_t}
\frac{1}{|\mathcal{H}_t|}\sum_{y \in \mathcal{H}_t}
\mathbf{1}[a(y) \le a(x)].
\]
Higher $q_t$ means that the surrogate is selecting references from the top of
the observed score distribution.
We measure exploration by comparing the selected references with the current
top-history set $\mathcal{T}_t \subset \mathcal{H}_t$, defined as the
highest-scoring historical molecules up to a fixed cutoff:
\[
d_t =
\frac{1}{|\mathcal{S}_t|\,|\mathcal{T}_t|}
\sum_{x \in \mathcal{S}_t}\sum_{y \in \mathcal{T}_t}
\bigl(1-\mathrm{Tan}(x,y)\bigr),
\]
where $\mathrm{Tan}(x,y)$ is the Tanimoto similarity between ECFP4
fingerprints.
Larger $d_t$ indicates broader exploration, while decreasing $d_t$ together
with increasing $q_t$ indicates a transition toward exploitation.

\paragraph{Latent-space visualization.}
Each valid molecule is represented by a 2048-bit ECFP4 fingerprint.
To obtain a stable two-dimensional visualization, we first reduce the
fingerprint vectors with PCA and then apply t-SNE to the reduced
representation.
For each task, all methods are embedded in the same two-dimensional space so
that their spatial distributions are directly comparable.
To highlight promising regions, we pool the scores of all methods for that task
and define a global high-score threshold $\theta_q$ as the empirical
$q$-quantile of the pooled score distribution.
The highlighted set is then
\[
\mathcal{G}_q = \{x : a(x) \ge \theta_q \}.
\]
Because the same threshold is used for all methods, the highlighted points identify the same top portion of the score distribution in every panel.
The latent-space figure therefore, visualizes not only where each method
generates molecules, but also how much of that generation budget falls into
regions enriched with high-scoring candidates.

\section{Additional Results}
\label{app:results}

\subsection{Additional Convergence Figures}
\label{app:extra_convergence}

\begin{figure*}[ht]
    \centering
    \subfigure[BBBP Convergence]{
        \includegraphics[width=0.23\linewidth]{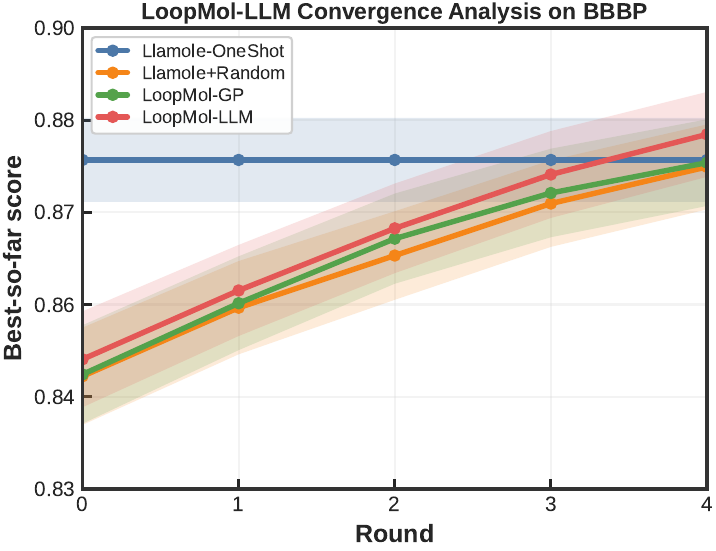}
    }
    \subfigure[BBBP Threshold (0.8)]{
        \includegraphics[width=0.23\linewidth]{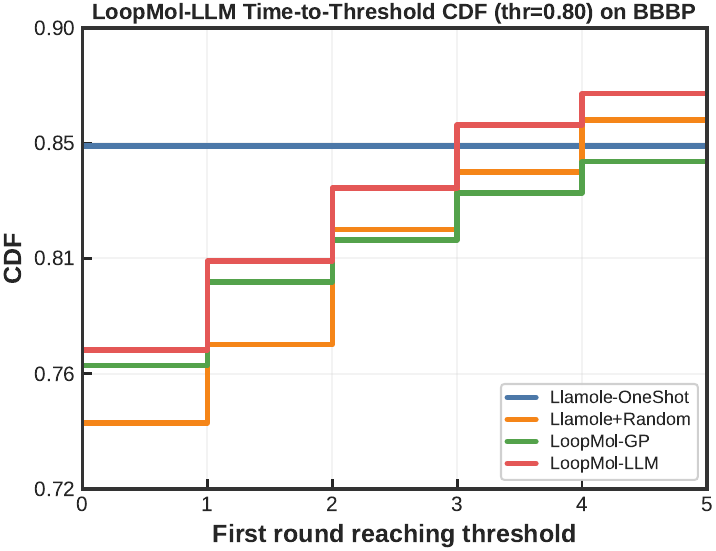}    }
    \subfigure[O$_2$ Convergence]{
        \includegraphics[width=0.23\linewidth]{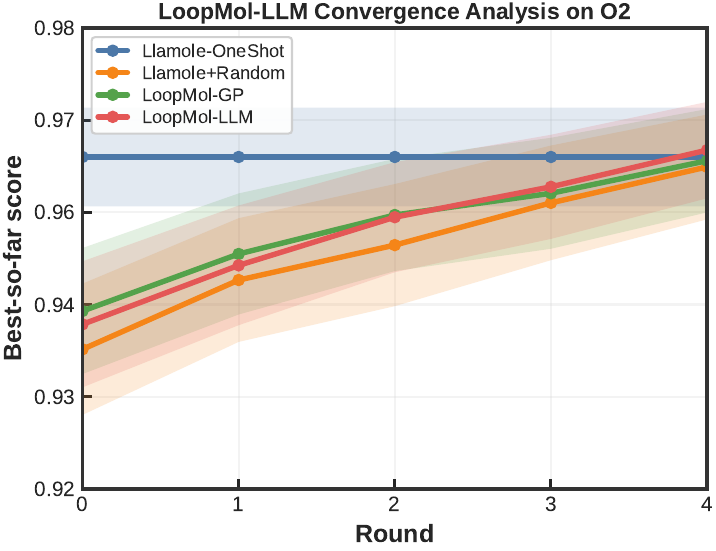}
    }
    \subfigure[O$_2$ Threshold (0.7)]{
        \includegraphics[width=0.23\linewidth]{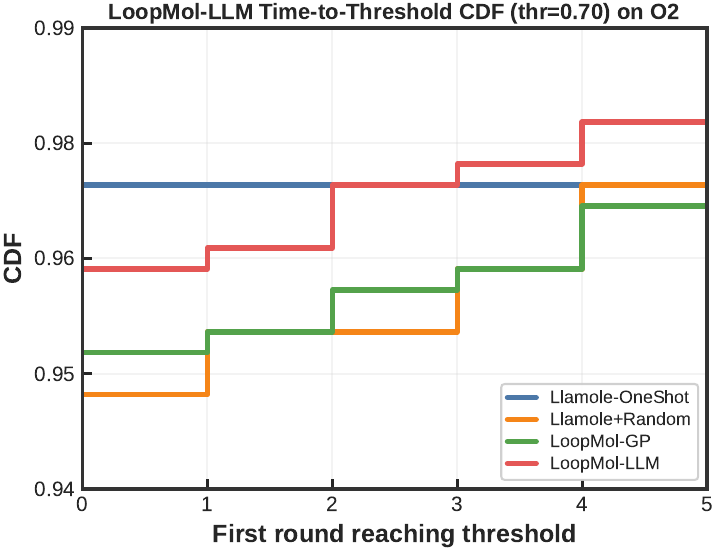}
    }
    \caption{Convergence diagnostics on BBBP and O$_2$.
    Left: mean best-so-far score as a function of round, with 95\% confidence
    intervals across test instances.
    Right: cumulative fraction of samples whose best-so-far score has crossed a
    fixed threshold by each round. In both tasks, iterative methods improve with
    rounds.}
    \label{fig_app:convergence}
\end{figure*}

Beyond the representative BACE and N$_2$ examples shown in the main text, we also include the corresponding convergence plots for additional tasks.
Figure \ref{fig_app:convergence} serves as a task-level complement to the main-text examples and
shows that the same trajectory-level improvement pattern holds broadly
across the benchmark.

\subsection{Visualizations of Generated Molecules}
\label{app:extra_latent}

\begin{figure*}[ht]
\centering
\includegraphics[width=\textwidth]{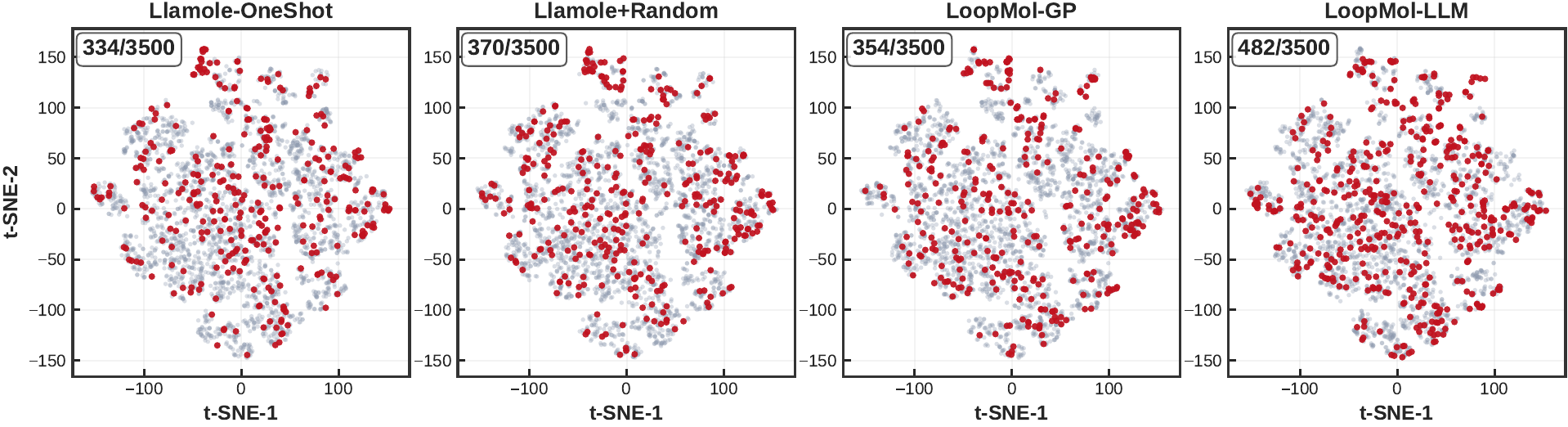}
\caption{Visualization of generated molecules on N$_2$, with high-scoring
candidates highlighted under a uniform threshold across methods. \method concentrates more of its search budget in regions enriched
with high-scoring candidates, without collapsing to a single cluster.}

\label{fig:tsne}\vspace{-3.9mm}
\end{figure*}

\begin{figure*}[ht]
\centering
\includegraphics[width=\textwidth]{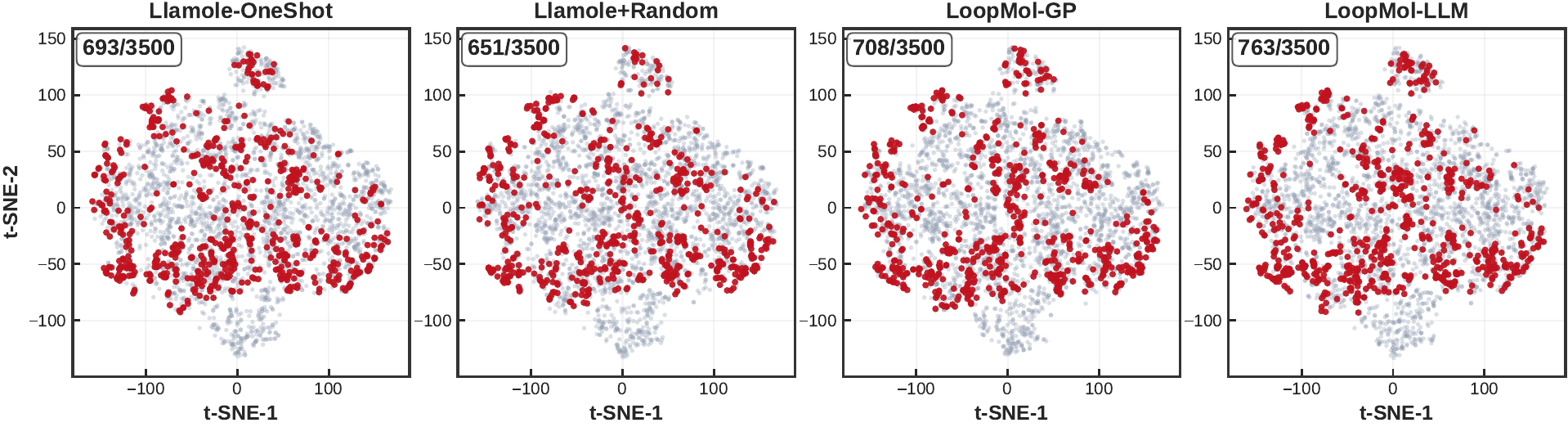}
\caption{Latent-space visualization of generated molecules on BACE.
Molecules are represented by ECFP4 fingerprints, projected to two dimensions,
and colored by whether they exceed a global high-score threshold shared across
methods. The numerator shown in each panel counts highlighted molecules, and
the denominator counts all displayed molecules for that method. A higher
highlighted fraction and a tighter concentration around promising regions
indicate more effective allocation of the search budget.}
\label{fig_app:tsne}
\end{figure*}

Figures~\ref{fig:tsne} and~\ref{fig_app:tsne} show t-SNE visualizations of
generated molecules on N$_2$ and BACE with the Llama backbone.
Molecules are represented by ECFP4 fingerprints and projected to two
dimensions using the protocol in Appendix~\ref{app:analysis_details}.
Within each task, a shared score threshold identifies high-scoring candidates
across methods.

\subsection{N$_2$ Case Study}
\label{app:n2_case}

We examine an N$_2$ optimization trace using the Llama backbone.
For this sample, the best score improves from $0.620$ at round 0 to $0.866$ by
round $3$.
The surrogate response describes structural patterns in the history and gives
reasons for the selected references, providing a human-readable account of
the search.
The full surrogate prompt template is given in Appendix~\ref{app:prompts}; here
we show a condensed version of the actual round-2 prompt and the corresponding
response.

\begin{AIbox}[colback=blue!2!white,colframe=blue!45!black,colbacktitle=blue!70!black]{Case Study: N$_2$ Optimization Trace}
\small
\textbf{Surrogate input(round 2).}
\begin{lstlisting}[style=promptstyle]
Task: Can you design a polymer with zero nitrogen or N2 gas permeability,
a fractional free volume of 0.415, and a specific molecular structure?
Target properties: N2=0.0, FFV=0.415, SC=3.33, SA=5.38
=== Optimization History ({n} molecules) ===
[10] *CC(B)(CCC(F)(F)F)C(=O)C(=O)CNC(=O)O* | score=0.6200 (N2:2.01)
[36] *CC(*)(C)C(=O)OCCC(F)(F)C(F)C(F)(F)C(F)F | score=0.5805 (N2:2.50)
...
=== Selection Principle (Bayesian Optimization) ===
...
\end{lstlisting}

\textbf{Surrogate output(round 2).}
\begin{lstlisting}[style=promptstyle]
ANALYSIS: After analyzing the optimization history, we observe that high-scoring 
molecules (e.g., [10], [36], [39]) tend to have a higher presence of heteroatoms (e.g., 
O, N, S, B, F, Cl) and a more complex molecular structure. They also often feature a
combination of polar and non-polar functional groups, which may contribute to their low 
N2 permeability. On the other hand, low-scoring molecules (e.g., [3], [6], [30]) often 
contain heavy metals (e.g., Na, Ge, Sn, Se) and/or unusual bonding patterns, which may 
hinder their performance. Unexplored directions that might be promising include 
incorporating more fluorine atoms, using alternative heteroatoms (e.g., P, As), or 
introducing new functional groups that can enhance the polymer's properties without 
increasing N2 permeability.
SELECTED: 10, 36, 39

GUIDE_FOCUS:
Refine high-scoring fluorinated structures and explore alternative heteroatoms to 
improve the property balance.
\end{lstlisting}
\end{AIbox}

In this response, the surrogate selects molecules 10, 36, and 39 and recommends
refining fluorinated structures. The selected references and the guidance can
be read alongside the model's analysis and the observed scores.

\subsection{BACE Hard Case}
\label{app:extra_cases}

\begin{figure}[!b]
    \centering
\includegraphics[width=0.7\linewidth]{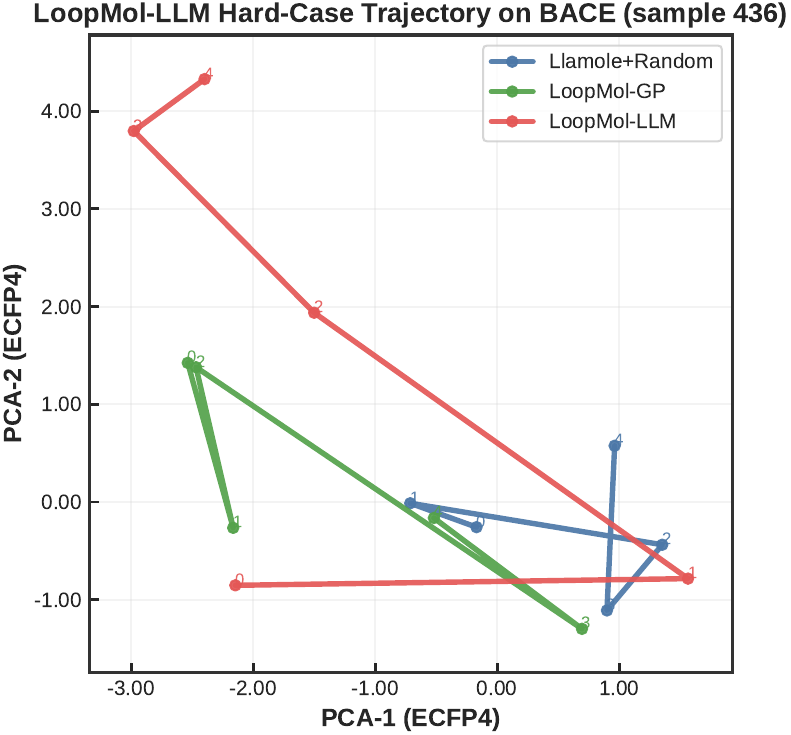}
    \caption{Latent trajectory of a hard case on BACE (sample 436).
    Each point is the best-scoring molecule found in each round, projected by PCA over
    ECFP4 fingerprints. Compared with the smoother trajectories of easier examples, the \methodllm path for this sample makes large early jumps and only reaches a stronger region in the last two rounds.}
    \label{fig:hard_case_436}
\end{figure}

We complement the N$_2$ example with a hard case from the Llama backbone on
BACE and show the full iterative trace.
This example is informative because the optimization does not improve smoothly:
the best round-level score drops from $0.4767$ at warm start to $0.4300$ and
$0.4140$ in rounds 1 and 2, then recovers only in the later rounds to
$0.5700$ and $0.6208$.
The case, therefore, illustrates a failure-prone regime in which effective refinement emerges only after several non-monotonic updates.

The warm-start round 0 does not involve any surrogate call; it initializes the history that is
later consumed by the surrogate.
We then provide the full surrogate interaction trace for rounds 1--4.
The box below contains the actual prompt passed to the surrogate and the
corresponding model output for this sample.
Since BACE uses the top-k only interface,
the surrogate output here contains \texttt{ANALYSIS} and \texttt{SELECTED}
fields only, without an additional guidance sentence.
Several aspects of this trace are notable.
First, the surrogate remains semantically coherent across rounds: it repeatedly
emphasizes aromatic rings, heterocycles, halogen substitution, and polar
functional groups as candidate design cues for BACE.
Second, these consistent local heuristics do not immediately translate into
monotonic optimization progress.
The trajectory initially moves away from the best warm-start point and only later returns to a stronger region, which is shown in
Figure~\ref{fig:hard_case_436}.

\begin{AIbox}{Full Iteration Trace for Hard Case 436}
\small
\textbf{Round 0: Warm Start}
\begin{lstlisting}[style=promptstyle]
Round 0 is warm start, so no surrogate call is made.
The generator receives only the original MolQA instruction.

Top warm-start molecules:
[1] N#Cc1cnc(S)c(O)c1 | score=0.4767
[2] \nolinkurl{CC(C)(C)OC(=O)N1CCc2ccccc2C(=O)N1c1ccc(C)cc1C(=O)N1CCc2ccccc2C(=O)N1 | score=0.3700
[3] O=C(OC(F)C(F)c1c(F)c(F)c(F)c(F)c1C(F)(F)F)C(F)(F)C(F)(F)C(F)F | score=0.3400
[4] O=[N+](O)c1cc(F)c(F)c(F)c1C(F)(F)C(F)(F)F | score=0.3300
[5] O=C(O)CC(O)(c1cc(F)c(F)cc1F)C(F)(F)F | score=0.3250
\end{lstlisting}

\medskip
\textbf{Round 1: Surrogate Prompt}
\begin{lstlisting}[style=promptstyle]
You are helping optimize molecular design. Your job is to select 3 reference molecules 
from the history below. These will guide the next round of molecule generation.

=== Task ===
Can you design a molecule that inhibits Human Immunodeficiency Virus and Beta-Secretase 
1, with a molecular weight of 646.76, four aromatic rings, and 13 rotatable bonds, 
and describe its synthesis?
Target properties: HIV=1.0, BACE=1.0, SC=2.11, SA=3.09
Property ranges:
  HIV: target=1.0, dataset range=[0.0, 1.0]
  BACE: target=1.0, dataset range=[0.0, 1.0]
  SC: target=2.11, dataset range=[1.0, 5.0]
  SA: target=3.09, dataset range=[1.0, 8.48]

=== Optimization History (30 molecules) ===
[1] CC=CC(=O)OC(CN)c1c(F)c(F)c(F)c(F)c1F | score=0.1800 (BACE:0.18)
[2] O=C(OC(F)(F)F)C(=O)C(=O)c1c(F)c(F)c(F)c(F)c1[N+](=O)O | score=0.2300 (BACE:0.23)
[3] COC(=O)Cc1ccc(Nc2cc(C)cc(-c3ccccc3)c2)cc1 | score=0.2467 (BACE:0.25)
[4] FC1=C(F)C(C(F)(F)C(F)(F)F)=C(F)C(F)=C(F)C(C(F)(F)C(F)(F)C(F)(F)F)=C1F | score=0.1100 (BACE:0.11)
[5] O=C(OC(F)C(F)c1c(F)c(F)c(F)c(F)c1C(F)(F)F)C(F)(F)C(F)(F)C(F)F | score=0.3400 (BACE:0.34)
[6] O=C(O)CC(O)(c1cc(F)c(F)cc1F)C(F)(F)F | score=0.3250 (BACE:0.32)
[7] Fc1c(F)c(F)c(F)c(F)c(F)c(F)c(F)c(F)c(F)c(F)c(F)c(F)c(F)c(F)c(F)c(F)c1F | score=0.0900 (BACE:0.09)
[8] COc1cc(OC[Si](C)(C)C)cc(Br)c1C=O | score=0.2100 (BACE:0.21)
[9] CCOC(=O)c1ccc(-c2ccc(-c3ccc(O)cc3)cc2)cc1 | score=0.2900 (BACE:0.29)
[10] CCOC(=O)c1ccc(OCc2ccc3c(c2)C(=O)Nc1=6c(n2ccc(C#N)cc2)c(OC)c(OC)cc6)cc2ccc
(C=7cnnn[nH]7)cc2 | score=0.0000
[11] CC(C)(C)OC(=O)N1CCc2ccccc2C(=O)N1c1ccc(C)cc1C(=O)N1CCc2ccccc2C(=O)N1 | score=0.3700 (BACE:0.37)
[12] CCOC(=O)c1c(Nc2nc3ccccc3c2)c(C)N1c1ccc(C(=O)OC)cc1 | score=0.0000
[13] Fc1cc(F)c(OC(F)(F)C(F)(F)C(F)(F)F)c(F)c1 | score=0.2990 (BACE:0.30)
[14] O=[N+](O)c1cc(C(F)(F)C(F)(F)C(F)(F)F)c(F)c(F)c1[N+](=O)O | score=0.2200 (BACE:0.22)
[15] FC1=C(F)C(F)=C(F)C(F)=C(F)C(F)=C(Br)C(F)=C(C(F)(F)F)C(F)=C(F)C(F)=C(F)C(F)
=C1F | score=0.1000 (BACE:0.10)
[16] N#Cc1cnc(S)c(O)c1 | score=0.4767 (BACE:0.48)
[17] O=[N+](O)c1cc(F)c(F)c(F)c1C(F)(F)C(F)(F)F | score=0.3300 (BACE:0.33)
[18] CCOC(=O)c1ccc(C)c(C)c(-c2nc3cc(C)cc(C)c(-c3nc2=O)n3cc(C)cc(C)c(C)c3c[nH]n3)c1
| score=0.0000
[19] CC1(C)CCCN(c2ncccn2)c1C(=O)Nc1ccc(C(=O)Nc2ccc(C(=O)Nc3ccc(C(=O)Nc4ccc(C(=O)
Nc5ccc(C(=O)Nc1ccc(C(=O)N)cc1)cc5)cc4)cc3)cc1 | score=0.0000
[20] CC(C)(C)OC(=O)N[C@H]1CCCN(c2ccc3c(c2)C(=O)Nc2ccc(OC)cc2c3)CC1 | score=0.0000
[21] CC(C)CCNc1nc(Cc2ccccc2)c(=O)n1Cc1ccc(C)c(C(=O)O)cc1 | score=0.0000
[22] COC(=O)c1ccc(-c2c(-c3ccccc3)cc2)cc1 | score=0.0400 (BACE:0.04)
[23] Fc1cc(Cl)c(S)nc1Cl | score=0.2500 (BACE:0.25)
[24] COC(=O)C(C)(F)Oc1c(F)c(F)c(F)c(F)c(F)c(F)c(OC)c(F)c1F | score=0.2500 (BACE:0.25)
[25] CCOC(=O)C(C)C1=C(/C=C(\C)C)SC1C(=O)Nc1ccc(-c2c[nH]c3ccccc23)cc1 
| score=0.3100 (BACE:0.31)
[26] Fc1c(F)c(F)c(OC(F)(F)COC(F)(F)F)c(F)c1F | score=0.1840 (BACE:0.18)
[27] CCOC(=N)C(=S)Nc1c(F)c(F)c(F)c(F)c1F | score=0.2300 (BACE:0.23)
[28] CCOC(=O)C(C)C(=O)Nc1cccc(-c2cc3c(c2)C(=O)Nc4ccccc4C(=O)Nc2cc(Cl)ccc2C(=O)
Nc1=2)C(=O)Nc1ccccc1C(F)(F)F | score=0.0000
[29] FC1=C(F)C(F)=C(F)C(F)(F)C(F)=C(F)C(OC(F)(F)C(F)(F)F)=C1F | score=0.1400 (BACE:0.14)
[30] CCOC(=O)C=C(/C=C/C=C/B1=5\C(=O)Oc1ccc(O)cc1O)N1CCN(c2ccc3nccnc3c2)CC1 | score=0.0000

=== Current State ===
Best score: 0.4767 (molecule [16])
Round: 1/5
Stage: Early stage - favor more EXPLORATION (try diverse structures)

=== Selection Principle (Bayesian Optimization) ===
Do NOT simply pick the 3 highest-scoring molecules. Instead, balance:
- EXPLOIT: pick molecules with high scores - these are good structures to refine
- EXPLORE: pick molecules with diverse or unusual structures - even if scores are 
moderate, they may represent promising unexplored directions that could lead to 
breakthroughs.

Think of it like an investment portfolio: some safe bets (high-scoring) + some wild 
cards (diverse structures). This is how Bayesian Optimization balances exploitation 
and exploration to avoid getting stuck in local optima.

=== Output Format ===
First, write your analysis, then write your selection in this exact format:
ANALYSIS: (describe what structural patterns you observe in high-scoring vs low-scoring
molecules, and what unexplored directions might be promising).
SELECTED: [3 molecule numbers, comma-separated, e.g. 1,5,12,3,8]
\end{lstlisting}
\medskip
\textbf{Round 1: Surrogate Output}
\begin{lstlisting}[style=promptstyle]
ANALYSIS: 
High-scoring molecules (e.g., 16, 11, 25) tend to have complex structures with multiple 
rings, nitrogen-containing groups, and a variety of functional groups such as amides,
ureas, and sulfonamides. These structures often have a high degree of molecular 
complexity, which may be beneficial for inhibiting the target proteins. 

Low-scoring molecules (e.g., 1, 7, 22) tend to have simpler structures with fewer rings 
and fewer nitrogen-containing groups. These structures may lack the complexity needed 
to effectively interact with the target proteins.

Unexplored directions that might be promising include incorporating more 
sulfur-containing groups, exploring different ring systems, and introducing more 
complex nitrogen-containing groups. Molecules with unusual structures, such as those 
with a high degree of branching or unusual functional groups, may also be 
worth investigating.

SELECTED: 16, 11, 25
\end{lstlisting}

\medskip
\textbf{Round 2: Surrogate Prompt}
\begin{lstlisting}[style=promptstyle]
You are helping optimize molecular design. Your job is to select 3 reference molecules 
from the history below. These will guide the next round of molecule generation.

=== Task ===
Can you design a molecule that inhibits Human Immunodeficiency Virus and Beta-Secretase 
1, with a molecular weight of 646.76, four aromatic rings, and 13 rotatable bonds, and 
describe its synthesis?
Target properties: HIV=1.0, BACE=1.0, SC=2.11, SA=3.09
Property ranges:
  HIV: target=1.0, dataset range=[0.0, 1.0]
  BACE: target=1.0, dataset range=[0.0, 1.0]
  SC: target=2.11, dataset range=[1.0, 5.0]
  SA: target=3.09, dataset range=[1.0, 8.48]


=== Optimization History (40 molecules) ===
[1] CC=CC(=O)OC(CN)c1c(F)c(F)c(F)c(F)c1F | score=0.1800 (BACE:0.18)
[2] O=C(OC(F)(F)F)C(=O)C(=O)c1c(F)c(F)c(F)c(F)c1[N+](=O)O | score=0.2300 (BACE:0.23)
[3] COC(=O)Cc1ccc(Nc2cc(C)cc(-c3ccccc3)c2)cc1 | score=0.2467 (BACE:0.25)
[4] FC1=C(F)C(C(F)(F)C(F)(F)F)=C(F)C(F)=C(F)C(C(F)(F)C(F)(F)C(F)(F)F)=C1F | score=0.1100 (BACE:0.11)
[5] O=C(OC(F)C(F)c1c(F)c(F)c(F)c(F)c1C(F)(F)F)C(F)(F)C(F)(F)C(F)F | score=0.3400 (BACE:0.34)
[6] O=C(O)CC(O)(c1cc(F)c(F)cc1F)C(F)(F)F | score=0.3250 (BACE:0.32)
[7] Fc1c(F)c(F)c(F)c(F)c(F)c(F)c(F)c(F)c(F)c(F)c(F)c(F)c(F)c(F)c(F)c(F)c1F | score=0.0900 (BACE:0.09)
[8] COc1cc(OC[Si](C)(C)C)cc(Br)c1C=O | score=0.2100 (BACE:0.21)
[9] CCOC(=O)c1ccc(-c2ccc(-c3ccc(O)cc3)cc2)cc1 | score=0.2900 (BACE:0.29)
[10] CCOC(=O)c1ccc(OCc2ccc3c(c2)C(=O)Nc1=6c(n2ccc(C#N)cc2)c(OC)c(OC)cc6)cc2ccc
(C=7cnnn[nH]7)cc2 | score=0.0000
[11] CC(C)(C)OC(=O)N1CCc2ccccc2C(=O)N1c1ccc(C)cc1C(=O)N1CCc2ccccc2C(=O)N1 | score=0.3700 (BACE:0.37)
[12] CCOC(=O)c1c(Nc2nc3ccccc3c2)c(C)N1c1ccc(C(=O)OC)cc1 | score=0.0000
[13] Fc1cc(F)c(OC(F)(F)C(F)(F)C(F)(F)F)c(F)c1 | score=0.2990 (BACE:0.30)
[14] O=[N+](O)c1cc(C(F)(F)C(F)(F)C(F)(F)F)c(F)c(F)c1[N+](=O)O | score=0.2200 (BACE:0.22)
[15] FC1=C(F)C(F)=C(F)C(F)=C(F)C(F)=C(Br)C(F)=C(C(F)(F)F)C(F)=C(F)C(F)=C(F)C(F)=C1F | score=0.1000 (BACE:0.10)
[16] N#Cc1cnc(S)c(O)c1 | score=0.4767 (BACE:0.48)
[17] O=[N+](O)c1cc(F)c(F)c(F)c1C(F)(F)C(F)(F)F | score=0.3300 (BACE:0.33)
[18] CCOC(=O)c1ccc(C)c(C)c(-c2nc3cc(C)cc(C)c(-c3nc2=O)n3cc(C)cc(C)c(C)c3c[nH]n3)c1 | score=0.0000
[19] CC1(C)CCCN(c2ncccn2)c1C(=O)Nc1ccc(C(=O)Nc2ccc(C(=O)Nc3ccc(C(=O)Nc4ccc(C(=O)Nc5ccc
(C(=O)Nc1ccc(C(=O)N)cc1)cc5)cc4)cc3)cc1 | score=0.0000
[20] CC(C)(C)OC(=O)N[C@H]1CCCN(c2ccc3c(c2)C(=O)Nc2ccc(OC)cc2c3)CC1 | score=0.0000
[21] CC(C)CCNc1nc(Cc2ccccc2)c(=O)n1Cc1ccc(C)c(C(=O)O)cc1 | score=0.0000
[22] COC(=O)c1ccc(-c2c(-c3ccccc3)cc2)cc1 | score=0.0400 (BACE:0.04)
[23] Fc1cc(Cl)c(S)nc1Cl | score=0.2500 (BACE:0.25)
[24] COC(=O)C(C)(F)Oc1c(F)c(F)c(F)c(F)c(F)c(F)c(OC)c(F)c1F | score=0.2500 (BACE:0.25)
[25] CCOC(=O)C(C)C1=C(/C=C(\C)C)SC1C(=O)Nc1ccc(-c2c[nH]c3ccccc23)cc1 | score=0.3100 (BACE:0.31)
[26] Fc1c(F)c(F)c(OC(F)(F)COC(F)(F)F)c(F)c1F | score=0.1840 (BACE:0.18)
[27] CCOC(=N)C(=S)Nc1c(F)c(F)c(F)c(F)c1F | score=0.2300 (BACE:0.23)
[28] CCOC(=O)C(C)C(=O)Nc1cccc(-c2cc3c(c2)C(=O)Nc4ccccc4C(=O)Nc2cc(Cl)ccc2C(=O)
Nc1=2)C(=O)Nc1ccccc1C(F)(F)F | score=0.0000
[29] FC1=C(F)C(F)=C(F)C(F)(F)C(F)=C(F)C(OC(F)(F)C(F)(F)F)=C1F | score=0.1400 (BACE:0.14)
[30] CCOC(=O)C=C(/C=C/C=C/B1=5\C(=O)Oc1ccc(O)cc1O)N1CCN(c2ccc3nccnc3c2)CC1 | score=0.0000
[31] Cc1ccc(-c2ccccc2)cc1C(=O)N1CCc2ccccc2C(=O)N1C(=O)N(C)CC(C)(C)OC(C)(C)C | score=0.3940 (BACE:0.39)
[32] Nc1nc(S)c(Sc2c(F)c(F)c(F)c(F)c2F)s1 | score=0.1900 (BACE:0.19)
[33] O=C(O)c1c(F)cc(CC(c2c(F)cc(F)c(F)c(F)c(F)c(F)c(F)c2F)C(F)(F)F)c(F)c1F | score=0.4300 (BACE:0.43)
[34] CC(C)C1=C(/C=C(\C)C)SC1C(=O)Nc1ccc(-c2c[nH]c3ccccc23)cc1 | score=0.2500 (BACE:0.25)
[35] Cc1c(-c2ccccc2)nc(NC(=O)c1-c1ccc(C)cc1)c1ccc(C)cc1 | score=0.0000
[36] COc1c(BOC(C)C)cc(C(F)(F)F)cc1N(O)O | score=0.2400 (BACE:0.24)
[37] O=C(C=CO)c1c(F)c(F)c(F)c(F)c1C(F)(F)F | score=0.1800 (BACE:0.18)
[38] O=CC1=C(F)C(F)=C(F)C(F)=C(F)C(F)=C(F)C(F)=C(F)C(F)=C1F | score=0.1700 (BACE:0.17)
[39] CCc1cc(C)c(C)c(-c2nc3ccccc3s2)cc1C(=O)N1CCc2ccccc2C(=O)N1 | score=0.0000
[40] O[SH](O)(O)(C1=C(F)C(F)=C(F)C(F)=C(F)C1(F)F)C(F)(F)C(F)(F)F | score=0.1600 (BACE:0.16)

=== Current State ===
Best score: 0.4767 (molecule [16])
Round: 2/5
Stage: Mid stage - BALANCE exploration and exploitation

=== Selection Principle (Bayesian Optimization) ===
Do NOT simply pick the 3 highest-scoring molecules. Instead, balance:
- EXPLOIT: pick molecules with high scores - these are good structures to refine
- EXPLORE: pick molecules with diverse or unusual structures - even if scores are 
moderate, they may represent promising unexplored directions that could lead 
to breakthroughs

Think of it like an investment portfolio: some safe bets (high-scoring) + some wild 
cards (diverse structures). This is how Bayesian Optimization balances exploitation and 
exploration to avoid getting stuck in local optima.

=== Output Format ===
First, write your analysis, then write your selection in this exact format:

ANALYSIS: (describe what structural patterns you observe in high-scoring vs low-scoring
molecules, and what unexplored directions might be promising)

SELECTED: [3 molecule numbers, comma-separated, e.g. 1,5,12,3,8]
\end{lstlisting}
\medskip
\textbf{Round 2: Surrogate Output}
\begin{lstlisting}[style=promptstyle]
ANALYSIS: 

Analyzing the optimization history, I observe that high-scoring molecules (scores above 
0.3) tend to have a combination of structural features, including:

1. Fluorine atoms: Many high-scoring molecules contain multiple fluorine atoms, 
which may contribute to their potency against both HIV and BACE.
2. Nitrogen-containing rings: Molecules with nitrogen-containing rings, such as 
pyridines or pyrimidines, tend to have higher scores.
3. Oxygen-containing groups: Oxygen-containing groups, such as carbonyls or alcohols, 
are also common in high-scoring molecules.
4. Aromatic rings: The presence of aromatic rings, particularly those with multiple 
fused rings, is a common feature among high-scoring molecules.

In contrast, low-scoring molecules often lack these features or have fewer fluorine 
atoms, nitrogen-containing rings, or oxygen-containing groups.

Unexplored directions that might be promising include:

1. Increasing the number of fluorine atoms: While many high-scoring molecules contain 
multiple fluorine atoms, there is still room for exploration in this area.
2. Incorporating more nitrogen-containing rings: Molecules with multiple 
nitrogen-containing rings or novel ring systems might lead to breakthroughs.
3. Exploring different oxygen-containing groups: Alternative oxygen-containing groups,
such as ethers or esters, could be used to enhance potency.

SELECTED: 16, 31, 33

These three molecules were chosen to balance exploitation and exploration:

1. **Molecule 16** (score: 0.4767) is the current best molecule, with a high score and 
a promising structure that includes multiple nitrogen-containing rings and fluorine 
atoms. This molecule serves as a good example for refinement.
2. **Molecule 31** (score: 0.3940) has a moderate score but an interesting structure
that includes multiple aromatic rings and a nitrogen-containing ring. This molecule 
represents a promising direction for further exploration.
3. **Molecule 33** (score: 0.4300) has a high score and a unique structure that includes 
multiple fluorine atoms and an oxygen-containing group. This molecule offers a novel 
combination of features that might lead to breakthroughs.
\end{lstlisting}

\medskip
\textbf{Round 3: Surrogate Prompt}
\begin{lstlisting}[style=promptstyle]
You are helping optimize molecular design. Your job is to select 3 reference molecules 
from the history below. These will guide the next round of molecule generation.

=== Task ===
Can you design a molecule that inhibits Human Immunodeficiency Virus and Beta-Secretase 
1, with a molecular weight of 646.76, four aromatic rings, and 13 rotatable bonds, and 
describe its synthesis?
Target properties: HIV=1.0, BACE=1.0, SC=2.11, SA=3.09
Property ranges:
  HIV: target=1.0, dataset range=[0.0, 1.0]
  BACE: target=1.0, dataset range=[0.0, 1.0]
  SC: target=2.11, dataset range=[1.0, 5.0]
  SA: target=3.09, dataset range=[1.0, 8.48]


=== Optimization History (50 molecules) ===
[1] CC=CC(=O)OC(CN)c1c(F)c(F)c(F)c(F)c1F | score=0.1800 (BACE:0.18)
[2] O=C(OC(F)(F)F)C(=O)C(=O)c1c(F)c(F)c(F)c(F)c1[N+](=O)O | score=0.2300 (BACE:0.23)
[3] COC(=O)Cc1ccc(Nc2cc(C)cc(-c3ccccc3)c2)cc1 | score=0.2467 (BACE:0.25)
[4] FC1=C(F)C(C(F)(F)C(F)(F)F)=C(F)C(F)=C(F)C(C(F)(F)C(F)(F)C(F)(F)F)=C1F | score=0.1100 (BACE:0.11)
[5] O=C(OC(F)C(F)c1c(F)c(F)c(F)c(F)c1C(F)(F)F)C(F)(F)C(F)(F)C(F)F | score=0.3400 (BACE:0.34)
[6] O=C(O)CC(O)(c1cc(F)c(F)cc1F)C(F)(F)F | score=0.3250 (BACE:0.32)
[7] Fc1c(F)c(F)c(F)c(F)c(F)c(F)c(F)c(F)c(F)c(F)c(F)c(F)c(F)c(F)c(F)c(F)c1F | score=0.0900 (BACE:0.09)
[8] COc1cc(OC[Si](C)(C)C)cc(Br)c1C=O | score=0.2100 (BACE:0.21)
[9] CCOC(=O)c1ccc(-c2ccc(-c3ccc(O)cc3)cc2)cc1 | score=0.2900 (BACE:0.29)
[10] CCOC(=O)c1ccc(OCc2ccc3c(c2)C(=O)Nc1=6c(n2ccc(C#N)cc2)c(OC)c(OC)cc6)cc2ccc
(C=7cnnn[nH]7)cc2 | score=0.0000
[11] CC(C)(C)OC(=O)N1CCc2ccccc2C(=O)N1c1ccc(C)cc1C(=O)N1CCc2ccccc2C(=O)N1 | score=0.3700 (BACE:0.37)
[12] CCOC(=O)c1c(Nc2nc3ccccc3c2)c(C)N1c1ccc(C(=O)OC)cc1 | score=0.0000
[13] Fc1cc(F)c(OC(F)(F)C(F)(F)C(F)(F)F)c(F)c1 | score=0.2990 (BACE:0.30)
[14] O=[N+](O)c1cc(C(F)(F)C(F)(F)C(F)(F)F)c(F)c(F)c1[N+](=O)O | score=0.2200 (BACE:0.22)
[15] FC1=C(F)C(F)=C(F)C(F)=C(F)C(F)=C(Br)C(F)=C(C(F)(F)F)C(F)=C(F)C(F)=C(F)C(F)=C1F 
| score=0.1000 (BACE:0.10)
[16] N#Cc1cnc(S)c(O)c1 | score=0.4767 (BACE:0.48)
[17] O=[N+](O)c1cc(F)c(F)c(F)c1C(F)(F)C(F)(F)F | score=0.3300 (BACE:0.33)
[18] CCOC(=O)c1ccc(C)c(C)c(-c2nc3cc(C)cc(C)c(-c3nc2=O)n3cc(C)cc(C)c(C)c3c[nH]n3)c1 | score=0.0000
[19] CC1(C)CCCN(c2ncccn2)c1C(=O)Nc1ccc(C(=O)Nc2ccc(C(=O)Nc3ccc(C(=O)Nc4ccc(C(=O)Nc5ccc
(C(=O)Nc1ccc(C(=O)N)cc1)cc5)cc4)cc3)cc1 | score=0.0000
[20] CC(C)(C)OC(=O)N[C@H]1CCCN(c2ccc3c(c2)C(=O)Nc2ccc(OC)cc2c3)CC1 | score=0.0000
[21] CC(C)CCNc1nc(Cc2ccccc2)c(=O)n1Cc1ccc(C)c(C(=O)O)cc1 | score=0.0000
[22] COC(=O)c1ccc(-c2c(-c3ccccc3)cc2)cc1 | score=0.0400 (BACE:0.04)
[23] Fc1cc(Cl)c(S)nc1Cl | score=0.2500 (BACE:0.25)
[24] COC(=O)C(C)(F)Oc1c(F)c(F)c(F)c(F)c(F)c(F)c(OC)c(F)c1F | score=0.2500 (BACE:0.25)
[25] CCOC(=O)C(C)C1=C(/C=C(\C)C)SC1C(=O)Nc1ccc(-c2c[nH]c3ccccc23)cc1 | score=0.3100 (BACE:0.31)
[26] Fc1c(F)c(F)c(OC(F)(F)COC(F)(F)F)c(F)c1F | score=0.1840 (BACE:0.18)
[27] CCOC(=N)C(=S)Nc1c(F)c(F)c(F)c(F)c1F | score=0.2300 (BACE:0.23)
[28] CCOC(=O)C(C)C(=O)Nc1cccc(-c2cc3c(c2)C(=O)Nc4ccccc4C(=O)Nc2cc(Cl)ccc2C(=O)Nc1=2)
C(=O)Nc1ccccc1C(F)(F)F | score=0.0000
[29] FC1=C(F)C(F)=C(F)C(F)(F)C(F)=C(F)C(OC(F)(F)C(F)(F)F)=C1F | score=0.1400 (BACE:0.14)
[30] CCOC(=O)C=C(/C=C/C=C/B1=5\C(=O)Oc1ccc(O)cc1O)N1CCN(c2ccc3nccnc3c2)CC1 | score=0.0000
[31] Cc1ccc(-c2ccccc2)cc1C(=O)N1CCc2ccccc2C(=O)N1C(=O)N(C)CC(C)(C)OC(C)(C)C | score=0.3940 (BACE:0.39)
[32] Nc1nc(S)c(Sc2c(F)c(F)c(F)c(F)c2F)s1 | score=0.1900 (BACE:0.19)
[33] O=C(O)c1c(F)cc(CC(c2c(F)cc(F)c(F)c(F)c(F)c(F)c(F)c2F)C(F)(F)F)c(F)c1F | score=0.4300 (BACE:0.43)
[34] CC(C)C1=C(/C=C(\C)C)SC1C(=O)Nc1ccc(-c2c[nH]c3ccccc23)cc1 | score=0.2500 (BACE:0.25)
[35] Cc1c(-c2ccccc2)nc(NC(=O)c1-c1ccc(C)cc1)c1ccc(C)cc1 | score=0.0000
[36] COc1c(BOC(C)C)cc(C(F)(F)F)cc1N(O)O | score=0.2400 (BACE:0.24)
[37] O=C(C=CO)c1c(F)c(F)c(F)c(F)c1C(F)(F)F | score=0.1800 (BACE:0.18)
[38] O=CC1=C(F)C(F)=C(F)C(F)=C(F)C(F)=C(F)C(F)=C(F)C(F)=C1F | score=0.1700 (BACE:0.17)
[39] CCc1cc(C)c(C)c(-c2nc3ccccc3s2)cc1C(=O)N1CCc2ccccc2C(=O)N1 | score=0.0000
[40] O[SH](O)(O)(C1=C(F)C(F)=C(F)C(F)=C(F)C1(F)F)C(F)(F)C(F)(F)F | score=0.1600 (BACE:0.16)
[41] CC(C)(C)OC(=O)N1CCc2ccccc2C(=O)N1C(=O)N(C)CCc1ccc(-c2ccccc2)cc1C(=O)N1CCc2ccccc2C
(=O)N1C(=O)N(C)CC(C)(C)OC(C)(C)C | score=0.4140 (BACE:0.41)
[42] CC(C)(C)OC(=O)N1CCc2ccccc2C(=O)N1C(=O)Nc1cccc(C)c1C(F)(F)F | score=0.3200 (BACE:0.32)
[43] CC(C)(C)OC(=O)N1CCc2ccccc2C(=O)N1C(=O)Nc1ccc(C=8=8)cc1C(=O)N(C)CC(C)(C)OC(C)(C)C | score=0.0000
[44] O=[N+](O)c1c(F)c(F)c(F)c(F)c1[N+](=O)OF | score=0.1600 (BACE:0.16)
[45] FC1=C(F)C(F)=C(F)C(F)=C(F)C([Se]c2c(F)c(F)c(F)c(F)c2F)=C(F)C(F)=C(F)C(F)=C1F | score=0.1400 (BACE:0.14)
[46] O=[N+](O)C(c1cc(F)c(F)cc1F)C(F)(F)C(F)(F)F | score=0.2900 (BACE:0.29)
[47] O=S(=O)(OC(F)(F)C(F)(F)F)C(F)(F)c1c(F)c(F)c(F)c(F)c1F | score=0.1400 (BACE:0.14)
[48] CC(C)OC(=O)N1CCc2ccccc2C(=O)N1C(=O)Nc1cc(C=8=7)cc(C=8=2)cc1C(=O)Nc1cc(C=1=O)
ccc1C(=O)N1CCc2ccccc2C(=O)N1C(=O)N(C)CC(C)(C)OC(C)(C)C | score=0.0000
[49] CC(C)(C)OC(=O)N1CCc2ccccc2C(=O)N1C(=O)Nc1ccc(C#N)c(Cl)c1C(=O)O | score=0.3400 (BACE:0.34)
[50] CC(C)(C)OC(=O)N1CCc2ccccc2C(=O)N1C(=O)Nc1ccc(C(C)(C)OC(C)(C)C)cc1 | score=0.2267 (BACE:0.23)

=== Current State ===
Best score: 0.4767 (molecule [16])
Round: 3/5
Stage: Mid stage - BALANCE exploration and exploitation

=== Selection Principle (Bayesian Optimization) ===
Do NOT simply pick the 3 highest-scoring molecules. Instead, balance:
- EXPLOIT: pick molecules with high scores - these are good structures to refine
- EXPLORE: pick molecules with diverse or unusual structures - even if scores are
moderate, they may represent promising unexplored directions that could lead to
breakthroughs

Think of it like an investment portfolio: some safe bets (high-scoring) + some wild 
cards (diverse structures). This is how Bayesian Optimization balances exploitation and 
exploration to avoid getting stuck in local optima.

=== Output Format ===
First, write your analysis, then write your selection in this exact format:

ANALYSIS: (describe what structural patterns you observe in high-scoring vs low-scoring 
molecules, and what unexplored directions might be promising)

SELECTED: [3 molecule numbers, comma-separated, e.g. 1,5,12,3,8]
\end{lstlisting}
\medskip
\textbf{Round 3: Surrogate Output}
\begin{lstlisting}[style=promptstyle]
ANALYSIS: 

Upon analyzing the optimization history, we observe that high-scoring molecules (e.g., 
16, 11, 31, 33, 49) tend to have a combination of features such as:

1. Fluorine substitution: Many high-scoring molecules have multiple fluorine atoms
attached to aromatic rings, which may enhance their biological activity.
2. Nitrogen-containing heterocycles: Compounds with nitrogen-containing rings (e.g., 
pyridine, pyrimidine) often have high scores, suggesting their importance in the design.
3. Aromatic ring systems: Molecules with multiple aromatic rings, especially those with
specific substitution patterns, tend to perform well.
4. Polar functional groups: The presence of polar functional groups (e.g., -OH, -NH, 
-O-) is common in high-scoring molecules.

Low-scoring molecules, on the other hand, often lack these features or have different
substitution patterns.

Unexplored directions that might be promising include:

1. Incorporating sulfur or selenium atoms into the molecule, as seen in molecules 32 
and 47.
2. Exploring different nitrogen-containing heterocycles, such as triazines orazines.
3. Investigating the effect of different aromatic ring systems, such as benzothiophene 
or indole.
4. Introducing more polar functional groups, like amide or urea, to enhance binding 
affinity.

SELECTED: 16, 11, 49
\end{lstlisting}

\medskip
\textbf{Round 4: Surrogate Prompt}
\begin{lstlisting}[style=promptstyle]
You are helping optimize molecular design. Your job is to select 3 reference molecules 
from the history below. These will guide the next round of molecule generation.

=== Task ===
Can you design a molecule that inhibits Human Immunodeficiency Virus and Beta-Secretase 
1, with a molecular weight of 646.76, four aromatic rings, and 13 rotatable bonds, and
describe its synthesis?
Target properties: HIV=1.0, BACE=1.0, SC=2.11, SA=3.09
Property ranges:
  HIV: target=1.0, dataset range=[0.0, 1.0]
  BACE: target=1.0, dataset range=[0.0, 1.0]
  SC: target=2.11, dataset range=[1.0, 5.0]
  SA: target=3.09, dataset range=[1.0, 8.48]


=== Optimization History (60 molecules) ===
[1] CC=CC(=O)OC(CN)c1c(F)c(F)c(F)c(F)c1F | score=0.1800 (BACE:0.18)
[2] O=C(OC(F)(F)F)C(=O)C(=O)c1c(F)c(F)c(F)c(F)c1[N+](=O)O | score=0.2300 (BACE:0.23)
[3] COC(=O)Cc1ccc(Nc2cc(C)cc(-c3ccccc3)c2)cc1 | score=0.2467 (BACE:0.25)
[4] FC1=C(F)C(C(F)(F)C(F)(F)F)=C(F)C(F)=C(F)C(C(F)(F)C(F)(F)C(F)(F)F)=C1F | score=0.1100 (BACE:0.11)
[5] O=C(OC(F)C(F)c1c(F)c(F)c(F)c(F)c1C(F)(F)F)C(F)(F)C(F)(F)C(F)F | score=0.3400 (BACE:0.34)
[6] O=C(O)CC(O)(c1cc(F)c(F)cc1F)C(F)(F)F | score=0.3250 (BACE:0.32)
[7] Fc1c(F)c(F)c(F)c(F)c(F)c(F)c(F)c(F)c(F)c(F)c(F)c(F)c(F)c(F)c(F)c(F)c1F | score=0.0900 (BACE:0.09)
[8] COc1cc(OC[Si](C)(C)C)cc(Br)c1C=O | score=0.2100 (BACE:0.21)
[9] CCOC(=O)c1ccc(-c2ccc(-c3ccc(O)cc3)cc2)cc1 | score=0.2900 (BACE:0.29)
[10] CCOC(=O)c1ccc(OCc2ccc3c(c2)C(=O)Nc1=6c(n2ccc(C#N)cc2)c(OC)c(OC)cc6)cc2ccc
(C=7cnnn[nH]7)cc2 | score=0.0000
[11] CC(C)(C)OC(=O)N1CCc2ccccc2C(=O)N1c1ccc(C)cc1C(=O)N1CCc2ccccc2C(=O)N1 | score=0.3700 (BACE:0.37)
[12] CCOC(=O)c1c(Nc2nc3ccccc3c2)c(C)N1c1ccc(C(=O)OC)cc1 | score=0.0000
[13] Fc1cc(F)c(OC(F)(F)C(F)(F)C(F)(F)F)c(F)c1 | score=0.2990 (BACE:0.30)
[14] O=[N+](O)c1cc(C(F)(F)C(F)(F)C(F)(F)F)c(F)c(F)c1[N+](=O)O | score=0.2200 (BACE:0.22)
[15] FC1=C(F)C(F)=C(F)C(F)=C(F)C(F)=C(Br)C(F)=C(C(F)(F)F)C(F)=C(F)C(F)=C(F)C(F)=C1F | score=0.1000 (BACE:0.10)
[16] N#Cc1cnc(S)c(O)c1 | score=0.4767 (BACE:0.48)
[17] O=[N+](O)c1cc(F)c(F)c(F)c1C(F)(F)C(F)(F)F | score=0.3300 (BACE:0.33)
[18] CCOC(=O)c1ccc(C)c(C)c(-c2nc3cc(C)cc(C)c(-c3nc2=O)n3cc(C)cc(C)c(C)c3c[nH]n3)c1 | score=0.0000
[19] CC1(C)CCCN(c2ncccn2)c1C(=O)Nc1ccc(C(=O)Nc2ccc(C(=O)Nc3ccc(C(=O)Nc4ccc(C(=O)Nc5ccc
(C(=O)Nc1ccc(C(=O)N)cc1)cc5)cc4)cc3)cc1 | score=0.0000
[20] CC(C)(C)OC(=O)N[C@H]1CCCN(c2ccc3c(c2)C(=O)Nc2ccc(OC)cc2c3)CC1 | score=0.0000
[21] CC(C)CCNc1nc(Cc2ccccc2)c(=O)n1Cc1ccc(C)c(C(=O)O)cc1 | score=0.0000
[22] COC(=O)c1ccc(-c2c(-c3ccccc3)cc2)cc1 | score=0.0400 (BACE:0.04)
[23] Fc1cc(Cl)c(S)nc1Cl | score=0.2500 (BACE:0.25)
[24] COC(=O)C(C)(F)Oc1c(F)c(F)c(F)c(F)c(F)c(F)c(OC)c(F)c1F | score=0.2500 (BACE:0.25)
[25] CCOC(=O)C(C)C1=C(/C=C(\C)C)SC1C(=O)Nc1ccc(-c2c[nH]c3ccccc23)cc1 | score=0.3100 (BACE:0.31)
[26] Fc1c(F)c(F)c(OC(F)(F)COC(F)(F)F)c(F)c1F | score=0.1840 (BACE:0.18)
[27] CCOC(=N)C(=S)Nc1c(F)c(F)c(F)c(F)c1F | score=0.2300 (BACE:0.23)
[28] CCOC(=O)C(C)C(=O)Nc1cccc(-c2cc3c(c2)C(=O)Nc4ccccc4C(=O)Nc2cc(Cl)ccc2C(=O)
Nc1=2)C(=O)Nc1ccccc1C(F)(F)F | score=0.0000
[29] FC1=C(F)C(F)=C(F)C(F)(F)C(F)=C(F)C(OC(F)(F)C(F)(F)F)=C1F | score=0.1400 (BACE:0.14)
[30] CCOC(=O)C=C(/C=C/C=C/B1=5\C(=O)Oc1ccc(O)cc1O)N1CCN(c2ccc3nccnc3c2)CC1 | score=0.0000
[31] Cc1ccc(-c2ccccc2)cc1C(=O)N1CCc2ccccc2C(=O)N1C(=O)N(C)CC(C)(C)OC(C)(C)C | score=0.3940 (BACE:0.39)
[32] Nc1nc(S)c(Sc2c(F)c(F)c(F)c(F)c2F)s1 | score=0.1900 (BACE:0.19)
[33] O=C(O)c1c(F)cc(CC(c2c(F)cc(F)c(F)c(F)c(F)c(F)c(F)c2F)C(F)(F)F)c(F)c1F | score=0.4300 (BACE:0.43)
[34] CC(C)C1=C(/C=C(\C)C)SC1C(=O)Nc1ccc(-c2c[nH]c3ccccc23)cc1 | score=0.2500 (BACE:0.25)
[35] Cc1c(-c2ccccc2)nc(NC(=O)c1-c1ccc(C)cc1)c1ccc(C)cc1 | score=0.0000
[36] COc1c(BOC(C)C)cc(C(F)(F)F)cc1N(O)O | score=0.2400 (BACE:0.24)
[37] O=C(C=CO)c1c(F)c(F)c(F)c(F)c1C(F)(F)F | score=0.1800 (BACE:0.18)
[38] O=CC1=C(F)C(F)=C(F)C(F)=C(F)C(F)=C(F)C(F)=C(F)C(F)=C1F | score=0.1700 (BACE:0.17)
[39] CCc1cc(C)c(C)c(-c2nc3ccccc3s2)cc1C(=O)N1CCc2ccccc2C(=O)N1 | score=0.0000
[40] O[SH](O)(O)(C1=C(F)C(F)=C(F)C(F)=C(F)C1(F)F)C(F)(F)C(F)(F)F | score=0.1600 (BACE:0.16)
[41] CC(C)(C)OC(=O)N1CCc2ccccc2C(=O)N1C(=O)N(C)CCc1ccc(-c2ccccc2)cc1C(=O)N1CCc2ccccc2
C(=O)N1C(=O)N(C)CC(C)(C)OC(C)(C)C | score=0.4140 (BACE:0.41)
[42] CC(C)(C)OC(=O)N1CCc2ccccc2C(=O)N1C(=O)Nc1cccc(C)c1C(F)(F)F | score=0.3200 (BACE:0.32)
[43] CC(C)(C)OC(=O)N1CCc2ccccc2C(=O)N1C(=O)Nc1ccc(C=8=8)cc1C(=O)N(C)CC(C)(C)OC(C)(C)C | score=0.0000
[44] O=[N+](O)c1c(F)c(F)c(F)c(F)c1[N+](=O)OF | score=0.1600 (BACE:0.16)
[45] FC1=C(F)C(F)=C(F)C(F)=C(F)C([Se]c2c(F)c(F)c(F)c(F)c2F)=C(F)C(F)=C(F)C(F)=C1F | score=0.1400 (BACE:0.14)
[46] O=[N+](O)C(c1cc(F)c(F)cc1F)C(F)(F)C(F)(F)F | score=0.2900 (BACE:0.29)
[47] O=S(=O)(OC(F)(F)C(F)(F)F)C(F)(F)c1c(F)c(F)c(F)c(F)c1F | score=0.1400 (BACE:0.14)
[48] CC(C)OC(=O)N1CCc2ccccc2C(=O)N1C(=O)Nc1cc(C=8=7)cc(C=8=2)cc1C(=O)Nc1cc(C=1=O)
ccc1C(=O)N1CCc2ccccc2C(=O)N1C(=O)N(C)CC(C)(C)OC(C)(C)C | score=0.0000
[49] CC(C)(C)OC(=O)N1CCc2ccccc2C(=O)N1C(=O)Nc1ccc(C#N)c(Cl)c1C(=O)O | score=0.3400 (BACE:0.34)
[50] CC(C)(C)OC(=O)N1CCc2ccccc2C(=O)N1C(=O)Nc1ccc(C(C)(C)OC(C)(C)C)cc1 | score=0.2267 (BACE:0.23)
[51] Fc1c(F)c(F)c([Se]c2c(F)c(F)c(F)c(F)c2F)c(F)c1F | score=0.1500 (BACE:0.15)
[52] Cc1cnc(S)c(O)c1.Cc1c(C(=O)N2CCc3ccccc3C(=O)N2)cccc1C(=O)N1CCc2ccccc2C(=O)N1 | score=0.0000
[53] O=S(=O)(Cl)C(Cl)=Cc1c(F)cc(F)c(F)c1F | score=0.2550 (BACE:0.25)
[54] O=CC1=C(F)C(F)=C(F)C(F)=C(F)C(F)=C(F)C(F)=C(F)C(F)=C1F | score=0.1700 (BACE:0.17)
[55] O=C(NC(=S)C(F)(F)F)c1cc(F)c(F)c(F)c1F | score=0.3350 (BACE:0.34)
[56] O=C(C(C(F)(F)C(F)F)C(F)(F)C(F)(F)C(F)(F)C(F)(F)F)C(F)(F)F | score=0.2200 (BACE:0.22)
[57] CC(C)(C)OC(=O)N1CCc2ccc(-c3ccccc3-c4ccccc4)cc2C(=O)N1c1ccc(C)cc1
C(=O)N1CCc2ccccc2C(=O)N1 | score=0.3600 (BACE:0.36)
[58] CC(C)(C)OC(=O)N1CCc2ccccc2C(=O)N1C(=O)Nc1ccc(C#N)c(Cl)c1C(=O)O | score=0.3400 (BACE:0.34)
[59] CC(C)(C)OC(=O)N1CCc2ccccc2C(=O)N1C(=O)Nc1ccc(C#N)cc1-c1ccc(C(C)C)cc1C(=O)O | score=0.5700 (BACE:0.57)
[60] O=Cc1cc(SCl)c(F)c(F)c1F | score=0.2500 (BACE:0.25)

=== Current State ===
Best score: 0.5700 (molecule [59])
Round: 4/5
Stage: Late stage - favor more EXPLOITATION (refine best structures)

=== Selection Principle (Bayesian Optimization) ===
Do NOT simply pick the 3 highest-scoring molecules. Instead, balance:
- EXPLOIT: pick molecules with high scores - these are good structures to refine
- EXPLORE: pick molecules with diverse or unusual structures - even if scores are
moderate, they may represent promising unexplored directions that could lead to
breakthroughs

Think of it like an investment portfolio: some safe bets (high-scoring) + some wild 
cards (diverse structures). This is how Bayesian Optimization balances exploitation and 
exploration to avoid getting stuck in local optima.

=== Output Format ===
First, write your analysis, then write your selection in this exact format:

ANALYSIS: (describe what structural patterns you observe in high-scoring vs low-scoring
molecules, and what unexplored directions might be promising)

SELECTED: [3 molecule numbers, comma-separated, e.g. 1,5,12,3,8]
\end{lstlisting}
\medskip
\textbf{Round 4: Surrogate Output}
\begin{lstlisting}[style=promptstyle]
ANALYSIS:

After analyzing the optimization history, I observe that high-scoring molecules (e.g., 
59, 57, 41) tend to have a combination of the following structural patterns:

1.  Multiple aromatic rings (often with multiple fused rings)
2.  Nitrogen-containing heterocycles (e.g., pyridines, pyrazines)
3.  Oxygen-containing functional groups (e.g., carbonyls, ethers)
4.  Halogen atoms (e.g., fluorine, chlorine) often attached to aromatic rings
5.  Sulfur-containing functional groups (e.g., sulfones, thioethers)

Low-scoring molecules (e.g., 10, 12, 20) tend to have fewer aromatic rings and fewer
heterocycles. They often have more aliphatic chains and fewer halogen atoms.

Unexplored directions that might be promising include:

1.  Incorporating more sulfur-containing functional groups (e.g., thioamides, thioureas)
2.  Exploring different types of heterocycles (e.g., imidazoles, triazoles)
3.  Increasing the number of aromatic rings and heterocycles in molecules with moderate
scores (e.g., 26, 30)

SELECTED: 59, 57, 41
\end{lstlisting}
\end{AIbox}

\subsection{Impact of Iteration Rounds}
\label{app:round_impact}

Table~\ref{tab:main} in the main text reports \method results with 5 optimization
rounds. To examine how the number of rounds affects performance, we conduct an
additional study on the Llama backbone using two representative tasks,
\textsc{BACE} and \textsc{O$_2$}. We vary the number of optimization rounds in
\(\{3,5,7,9,11\}\), while keeping all other settings fixed. In each round, 10 molecules are generated, and we compare against
\texttt{Llamole-OneShot} under the same evaluation protocol.

\begin{figure}[h]
    \centering
    \includegraphics[width=1\linewidth]{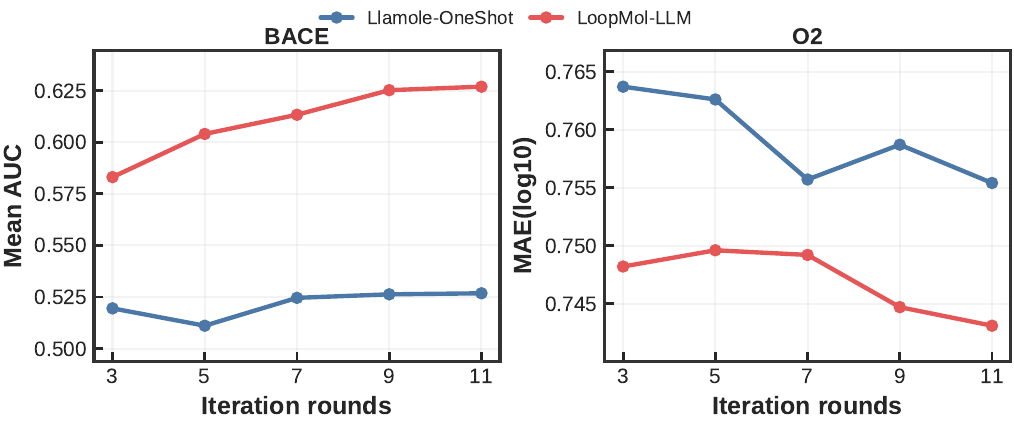}
    \caption{Effect of optimization round count on \texttt{Llamole-OneShot} and
    \method (Llama backbone). Left: \textsc{BACE}; right: \textsc{O$_2$}.
    All settings are fixed except the number of rounds
    (\(3,5,7,9,11\)). Each round generates 10 molecules.}
    \label{fig:round_impact_bace_o2}
\end{figure}

Figure~\ref{fig:round_impact_bace_o2} shows the resulting trends. On
\textsc{BACE}, where higher AUC is better, \method improves from 0.5831 at
3 rounds to 0.6269 at 11 rounds, while \texttt{Llamole-OneShot} remains around
0.51--0.53. On \textsc{O$_2$}, where lower \(\mathrm{MAE}(\log_{10})\) is better,
\method achieves lower error than \texttt{Llamole-OneShot} for all tested round
counts, with the error changing from 0.7482 at 3 rounds to 0.7431 at 11 rounds.
These results show that increasing the number of closed-loop rounds can improve
performance under the fixed per-round generation budget. Across all tested round counts, \method consistently outperforms
\texttt{Llamole-OneShot} on both tasks, showing that the closed-loop procedure
maintains an advantage over one-shot generation as the number of optimization
rounds increases.

\end{document}